\documentclass[pdflatex,sn-mathphys-ay]{sn-jnl}
\usepackage{array} 
\usepackage{amsthm}
\usepackage{fix-cm}
\usepackage{xcolor}
\usepackage{mathrsfs} 
\usepackage{makecell}
\usepackage{graphicx}
\usepackage{multirow}
\usepackage{tabularx}
\usepackage{textcomp}
\usepackage{manyfoot}
\usepackage{booktabs}
\usepackage{listings}
\usepackage{algorithm}
\usepackage{algorithmicx}%
\usepackage{algpseudocode}
\usepackage{threeparttable}
\usepackage[title]{appendix}
\usepackage{amsmath,amssymb,amsfonts,tcolorbox}
\newcolumntype{C}[1]{>{\centering\arraybackslash}p{#1}}   

\theoremstyle{thmstyleone}
\theoremstyle{thmstyletwo}
\theoremstyle{thmstylethree}

\begin{document}
\title[Article Title]{Linear Causal Discovery with Interventional Constraints}
\author*[1]{\fnm{Zhigao} \sur{Guo}}\email{zhigao.guo@strath.ac.uk}
\author[1]{\fnm{Feng} \sur{Dong}}\email{feng.dong@strath.ac.uk}
\affil*[1]{\orgdiv{Department of Computer and Information Sciences}, \orgname{University of Strathclyde}, \orgaddress{\country{UK}}}

\abstract{Incorporating causal knowledge and mechanisms is essential for refining causal models and improving downstream tasks, such as designing new treatments. In this paper, we introduce a novel concept in causal discovery, termed \textit{interventional constraints}, which differs fundamentally from interventional data. While interventional data require direct perturbations of variables, interventional constraints encode high-level causal knowledge in the form of inequality constraints on causal effects. For instance, in the Sachs dataset (\textcolor{blue}{\cite{Sachs-2005}}), Akt has been shown to be \textit{activated} by PIP3, meaning PIP3 exerts a \textit{positive} causal effect on Akt. Existing causal discovery methods allow enforcing structural constraints (e.g., requiring a causal path from PIP3 to Akt), but they may still produce incorrect causal conclusions, such as learning that ``PIP3 \textit{inhibits} Akt.'' Interventional constraints bridge this gap by explicitly constraining the total causal effect between variable pairs, ensuring learned models respect known causal influences. To formalize interventional constraints, we propose a metric to quantify total causal effects for \textit{linear} causal models and formulate the problem as a constrained optimization task, solved using a two-stage constrained optimization method. We evaluate our approach on real-world datasets and demonstrate that integrating interventional constraints not only improves model accuracy and ensures consistency with established findings, making models more explainable, but also facilitates the discovery of new causal relationships that would otherwise be costly to identify.}

\keywords{Causal discovery, Causal inference, Causal effect, Prior knowledge, Continuous optimization}

\maketitle

\section{Introduction}
\label{sec:intro}
Understanding causality is crucial for developing explainable, safe, fair, and robust machine learning models that generalize well to new environments (\textcolor{blue}{\cite{Pearl-seven-sparks, Causal-machine-learning-survey, Causal_machine_learning_nature}}). Causal discovery, an essential component of causality, reveals underlying causal mechanisms in data and provides insights into true causes and effects (\textcolor{blue}{\cite{peters-elements-2017,Like_DAGs,Kun-Survey-2023, Anthony-Survey-2023}}). This is particularly useful when experimental manipulation such as randomised trials is subject to limitations in costs, time and ethical restrictions (\textcolor{blue}{\cite{Causal_effect_nature}}). However, purely data-driven causal discovery methods often struggle with issues such as limited sample sizes, measurement bias, and noise. In many applications, human knowledge of known causal influences can be very useful to enhance the accuracy and  interpretability of causal discovery when integrated into learning (\textcolor{blue}{\cite{Anthony-knowledge-2023}}). 

While prior research has primarily focused on enforcing structural constraints to shape the causal graph, these methods do not constrain the causal effects (i.e., the parameters). In this paper, we introduce, interventional constraints, a previously unexplored category of high-level prior knowledge that simultaneously constrains both the causal structure and its associated causal effects. This is essential for improving downstream applications that rely on causal models. Our approach advances causal discovery by enabling more flexible, knowledge-guided inference while maintaining model interpretability and robustness. To illustrate the concept of interventional constraints, consider the widely used Sachs dataset (\textcolor{blue}{\cite{Sachs-2005}}) describing a signalling pathway in human immune cells. Biological experiments establish that PIP3 \textit{activates} Akt, meaning that PIP3 exerts a \textit{positive} causal effect on Akt. Such knowledge can serve as a testable constraint (\textcolor{blue}{\cite[p. 64]{pearl-causal-primer-2016}}) and be formulated as an interventional constraint. Hence, if a causal model predicts that PIP3 \textit{inhibits} Akt, it would violate the interventional constraint and contradict established evidence, even if the model includes a causal path from PIP3 to Akt. Importantly, such domain knowledge is prevalent across many fields. For instance, in epidemiology, it is well known that smoking \textit{increases} the risk of lung cancer; in economics, tax reductions often exert a \textit{positive} causal effect on consumer spending. Unlike fully experimental or interventional datasets that require directly perturbing variables, interventional constraints offer a way to incorporate such high-level causal information without the need for complete interventional data. This allows causal discovery to leverage high-level human knowledge as constraints, reducing reliance on accessing detailed, extensive experimental data. Hence, this newly proposed method offers a scalable and practical way to enhance causal discovery in many real-world settings. The main contributions of this paper are as follows:
\begin{itemize}
\item We introduce causal discovery with a new type of constraint, termed \textit{interventional constraints} to incorporate qualitative knowledge of causal effects into the learning process. Unlike existing constraints that mainly affect a model's structure, the interventional constraints regulate both the causal pathways (structure) and the causal effects (parameters) of the model.

\item We propose a metric that quantifies total causal effects between variable pairs in \textit{linear} causal models, capturing both direct and indirect effects, enabling the application of interventional constraints to causal pathways of any length.

\item We present a tailored two-stage mixed optimization approach to solve the problem of causal discovery with interventional constraints under the linear assumption.

\item We validate the proposed method on both synthetic and real-world data. Experiments on synthetic data demonstrate that interventional constraints are more effective than traditional path constraints. Real-world experiments further show that partial interventional constraints enable the identification of additional causal interactions (e.g., ``PKA \textit{inhibits} P38'') and causal paths (e.g., Mek $\to \dots \to$ Erk).
\end{itemize}

\noindent\textbf{Remark:} Within this paper, we focus on demonstrating causal discovery with interventional constraints in the \textit{linear} setting, the underlying concept of interventional constraints is general and can, in principle, be extended to nonlinear settings — a direction we identify as promising for future research. Hence this work serves as a preliminary step toward more general integrations of such knowledge. This is similar in spirit to the development of LiNGAM (\textcolor{blue}{\cite{LiNGAM}}) and NOTEARS (\textcolor{blue}{\cite{NOTEARS}}), which began with linear models and later inspired extensions to nonlinear frameworks. Our goal is to lay a foundation for future research extending interventional constraints to more complex, nonlinear scenarios.

\section{Related Work}
\label{sec:related_work}

Various approaches have been developed to integrate human or prior knowledge through structural constraints, including node ordering (e.g., $X_1 \prec X_3 \prec X_2$), edge constraints (e.g., $X_1 \to X_2$), path constraints (e.g., $X_1 \to \dots \to X_2$) and expert-provided structure information. Early methods, such as K2 algorithm \textcolor{blue}{\cite{K2}}, relied on predefined node ordering for Bayesian network structure learning. Subsequent works expanded on this by integrating multiple prior constraints, as seen in \textcolor{blue}{\cite{LiNGAM-prior}}, which enhanced LiNGAM-based causal discovery by incoporation of path constraints. More interactive approaches, such as those by \textcolor{blue}{\cite{meek1995causal}}, \textcolor{blue}{\cite{cano-method-2011}} and \textcolor{blue}{\cite{masegosa-interactive-2013}}, allowed for the incorporation of edges, path constraints and certain required edge orientations, enabling more flexible structure learning. Recent advancements have focused on refining structural priors and integrating domain knowledge in a more systematic manner. \textcolor{blue}{\cite{maximal-PDAG-2017}} proposed a method for incorporating edge orientations and partial ordering constraints into maximally oriented Partially Directed Acyclic Graphs (maximal PDAGs) learning, while \textcolor{blue}{\cite{Causal_knowledge_AISTATS_2020}} introduced tiered causal ordering into the FCI algorithm. \textcolor{blue}{\cite{KCRL}} utilized reinforcement learning to penalize edge constraint violations, thereby enforcing known causal relationships. Other works have leveraged approximate causal structures as priors. For instance, \textcolor{blue}{\cite{geffner-deep-2022}} utilized Completed Partially Directed Acyclic Graph (CPDAG) from the PC algorithm, while \textcolor{blue}{\cite{choo-advice-2023}} employed approximate DAGs obtained from expert input. In a more general framework, \textcolor{blue}{\cite{Anthony-knowledge-2023}} proposed integrating various structural priors into Bayesian network structure learning, demonstrating the impact of domain knowledge on causal structure learning. Their work aligns with efforts such as \textcolor{blue}{\cite{rittel-specifying-2023}}, who developed differentiable Bayesian models incorporating expert-specified edges and node ordering constraints. Several recent approaches incorporate edge constraints into continuous optimization frameworks. \textcolor{blue}{\cite{NTS-NOTEARS}} framed Dynamic Bayesian Network (DBN) structure learning as a continuous optimization problem incorporating edge constraints from One-Dimensional Convolutional Neural Networks (1D CNNs). Similarly, \textcolor{blue}{\cite{CAM-UV-prior}} integrated exclusion and temporal ordering constraints to improve causal additive model identification. \textcolor{blue}{\cite{Zidong-2024}} further extended this paradigm by integrating edge, path, and ordering constraints into differential causal discovery. Existing research on incorporating prior knowledge into causal discovery is summarized in Table~\ref{tab:prior_knowledge_summary}.
\begingroup
\begin{table*}[ht]
\footnotesize
\begin{tabularx}{\textwidth}{|>{\centering\arraybackslash}p{3.5cm} 
                                 |>{\centering\arraybackslash}p{3.5cm} 
                                 |>{\centering\arraybackslash}X|}
\hline
\textbf{Reference} & \textbf{Prior Type} & \textbf{Comments} \\
\hline

\textbf{\cite{K2}} 
 & Node ordering 
 & Pioneered predefined variable ordering for discrete Bayesian networks structure learning. \\
\hline

\textbf{\cite{meek1995causal}} 
 & Edge orientations 
 & Identifies causal relations shared by all DAGs consistent with data and background knowledge. \\
\hline

\textbf{\cite{LiNGAM-prior}} 
 & Path constraints 
 & Enhances LiNGAM with path constraints for improved linear causal structure identification. \\
\hline

\textbf{\cite{cano-method-2011}}, \textbf{\cite{masegosa-interactive-2013}}
 & Edge and path constraints
 & Enables interactive prior knowledge integration for structure learning. \\
\hline

\textbf{\cite{maximal-PDAG-2017}} 
 & Edge orientations, Markov equivalence, partial ordering 
 & Integrates prior to learn maximal PDAG. \\
\hline

\textbf{\cite{Causal_knowledge_AISTATS_2020}} 
 & Tiered causal ordering 
 & Integrates tiered causal ordering into FCI. \\
\hline

\textbf{\cite{KCRL}} 
 & Edge constraints 
 & Uses prior knowledge in reinforcement learning to penalize constraint-violating causal structures. \\
\hline

\textbf{\cite{geffner-deep-2022}} 
 & CP-DAG learned by the PC algorithm 
 & Leverages CP-DAG and domain knowledge to enhance causal recovery. \\
\hline

\textbf{\cite{rittel-specifying-2023}} 
 & Edge and ordering constraints
 & Refines DAG priors in a differentiable Bayesian framework to integrate expert-provided edges or node ordering constraints. \\
\hline

\textbf{\cite{Anthony-knowledge-2023}}
 & Various structural priors 
 & Integrates comprehensive structural priors into Bayesian network structure learning.\\
\hline

\textbf{\cite{choo-advice-2023}} 
 & Approximate DAG from experts 
 & Utilizes an approximate DAG as prior knowledge for robust causal structure recovery. \\
\hline

\textbf{\cite{NTS-NOTEARS}} 
 & Edge constraints 
 & Frames DBN structure learning as continuous optimization with edge constraints from 1D CNNs. \\
\hline

\textbf{\cite{CAM-UV-prior}} 
 & Exclusion and temporal ordering
 & Integrates prior knowledge to enhance causal additive model identification. \\
\hline

\textbf{\cite{Zidong-2024}} 
 & Edge, path and ordering constraints 
 & Incorporates edge, path, and ordering priors into differential causal discovery. \\
\hline
\end{tabularx}
\caption{Related work on incorporating prior knowledge in causal discovery}
\label{tab:prior_knowledge_summary}
\end{table*}
\endgroup

\section{Interventional Constraints}\label{sec:problem_setting}
This section introduces the novel concept of \textit{interventional constraints}, a new form of high-level causal knowledge that expresses the expected direction and strength of causal effects between variable pairs. We formally define these constraints and demonstrate how they can be incorporated into linear causal discovery, where causal effects are explicitly represented by edge weights and total effects along causal paths. 

\subsection{Definition}\label{sec:interventional_constraints}

\begin{tcolorbox}[colback=white, colframe=black, title=\textbf{Definition 3.1} (Interventional Constraints)]
Let $T_{i,j}$ be the total causal effect of variable $X_i$ on variable $X_j$. Interventional constraints specify whether this effect is positive or negative, such that $T_{i,j} > 0$ indicates a \textit{positive} effect, and $T_{i,j} < 0$ indicates a \textit{negative} effect.
\end{tcolorbox}

\noindent\textbf{Remark:} Note that our interventional constraints are qualitative and expressed as inequalities (e.g., $T_{i,j} > 0$), differing from the fine-grained quantitative interventional data. Unlike methods assuming direct experimental interventions (\textcolor{blue}{\cite{GIES, DCDI, Intervention_lippe, CD_intervention_Ke}}), our approach uses qualitative expert knowledge. Such constraints may originate not only from randomized controlled trials but also from broader domain evidence. For example, as Judea Pearl noted:  ``Consider the century-old debate concerning the effect of smoking on lung cancer. In 1964, the Surgeon General issued a report linking cigarette smoking to death, cancer, and most particularly lung cancer. The report was based on nonexperimental studies in which a strong correlation was found between smoking and lung cancer, and the claim was that the correlation found is causal: If we ban smoking, then the rate of cancer cases will be roughly the same as the one we find today among nonsmokers in the population.'' (\textcolor{blue}{\cite[p. 423]{pearl-causality-2009}}). This assertion can be represented as an interventional constraint in our framework, expressed as $T(\text{Smoking}, \text{Lung cancer}) > 0$. These constraints are significantly easier to specify compared to the detailed numerical values typically required in interventional datasets. Similarly, in the Sachs dataset (\textcolor{blue}{\cite{Sachs-2005}}), where prior biological knowledge indicates that PIP3 \emph{activates} Akt (i.e., $T(\text{PIP3}, \text{Akt}) > 0$) (\href{https://reactome.org/content/detail/R-HSA-1257604}{Reactome: R-HSA-1257604}), implying that PIP3 has a \emph{positive} causal effect on Akt. Traditional causal discovery might reveal a causal path from PIP3 to Akt but not guarantee its sign. In contrast, our method enforces consistency with such known effects without requiring detailed numerical interventional data. 

\subsection{Linear Causal Discovery with Interventional Constraints}

We consider causal discovery under the standard assumptions used in linear structural equation models:

\begin{itemize}
    \item Causal Sufficiency: All common causes of observed variables are included in the model, so there are no unmeasured confounders.
    \item Causal Markov Condition: Each variable is conditionally independent of its non-descendants given its parents, allowing the joint distribution to factorize according to the DAG.
    \item Faithfulness: All conditional independencies in the observed data correspond to d-separation relations in the true causal DAG.
    \item Linearity and Additive Gaussian Noise: Each variable is generated as a linear function of its parents, with an independent additive Gaussian noise term. The noise variances are assumed to be unequal or unknown.
\end{itemize}

In a linear causal model, each variable $X_i$ is a linear function of its direct causes $\text{Pa}(X_i)$ plus an independent additive noise term $z_i$:
\begin{equation}
X_i = \sum_{X_j \in \text{Pa}(X_i)} w_{ij} X_j + z_i, \quad i = 1, 2, \dots, d,
\end{equation}
where $w_{ij}$ denotes the direct causal effect of $X_j$ on $X_i$, and $z_i$ are mutually independent Gaussian noise terms with unequal (or unknown) variances. These weights form a weighted adjacency matrix $W \in \mathbb{R}^{d \times d}$, and the overall objective of causal discovery is to recover $W$) from observed data $X \in \mathbb{R}^{n \times d}$. We adopt the continuous optimization framework of NOTEARS (\textcolor{blue}{\cite{NOTEARS}}), where the estimation of $W$ is formulated as the following optimization problem:
\begin{equation}
\min_{W \in \mathbb{R}^{d \times d}} F(W)
\label{eq:obj_fun} 
\end{equation}
subject to
\begin{gather}
\delta_{ij} (T_{ij} - \delta_{ij}) > 0, \quad i \in \mathcal{C}, j \in \mathcal{T}, \label{eq:constraints} \\
\text{h}(W) = 0, \label{eq:h_w} 
\end{gather}
where the objective function is defined as
\begin{equation}
F(W) = \frac{1}{2n} \|X - XW\|_F^2 + \lambda \|W\|_1,
\label{eq:T_h_F}
\end{equation}
and the acyclicity constraint is imposed via
\begin{equation}
\text{h}(W) = \text{tr}\left(e^{W \circ W}\right) - d. \label{eq:h_w_expression}
\end{equation}
Here, the Frobenius norm penalizes prediction error, the $\ell_1$ norm encourages sparsity, and the exponential trace constraint enforces DAG-ness. The main addition beyond traditional causal discovery is the new interventional constraint in Equation~\ref{eq:constraints}, which encodes prior knowledge about causal effects through a lower-bound inequality on the total effect matrix $T$.
To encode expert knowledge, we impose:
\[
\delta_{ij}(T_{ij} - \delta_{ij}) > 0,
\]
which ensures that the total causal effect $T_{ij}$ exceeds threshold $\delta_{ij}$ in magnitude and matches its sign. For instance, if $\delta_{ij} = 0.1$, then $T_{ij} > 0.1$; if $\delta_{ij} = -0.1$, then $T_{ij} < -0.1$. The above constrained formulation is novel in jointly enforcing both acyclicity (via nonlinear equality) and interventional knowledge (via nonlinear inequality). Together, these constraints regulate both structure and parameters, distinguishing our method from prior work which only considers structural constraints.

\noindent\textbf{Remark:} For linear-Gaussian models with unequal (or unknown) noise variances, causal discovery is limited to identifying the Markov equivalence class (\textcolor{blue}{\cite{verma1990equivalence, LiNGAM, peters2014identifiability, Kun-Survey-2023}}). Introducing qualitative interventional constraints—expressed as inequality conditions on total causal effects—can help resolve causal directions by penalizing models that contradict known effect signs. However, we emphasize that the key novelty of our work does not lie in altering identifiability assumptions, but in proposing interventional constraints as a new form of knowledge-driven guidance, which directly imposes inequality constraints on total causal effects between variables.

For linear causal models, we have the following proposition to measure the total causal effect matrix below, which captures both direct and indirect causal effects between variables.

\begin{tcolorbox}[colback=white, colframe=black, title=\textbf{Proposition 3.1} (Total Causal Effects in Linear Models)]
In a linear causal model, the matrix $T$ encapsulates total causal effects (both direct and indirect) between variable pairs:
\begin{equation}
T = (I - W)^{-1} - I.
\label{eq:causal_effect_matrix}
\end{equation}
\end{tcolorbox}

\begin{proof}: 
In a linear causal model, each entry $w_{ij}$ represents the direct causal effect of variable $i$ on variable $j$ (\textcolor{blue}{\cite{pearl-causality-2009}}). The matrix $(I - W)^{-1}$ can be expanded as the series $I + W + W^2 + W^3 + \dots$, where higher powers of $W$ represent the effects of longer paths through the graph. For instance, $W$ captures the \textit{direct} causal effects between variables and $W^2$ represents the effects that pass through one intermediary variable (\textit{indirect} causal effects of length two). Subtracting the identity matrix $I$ from $(I - W)^{-1}$ removes the trivial self-effects of each variable, which are represented by the diagonal elements equal to 1 in $(I - W)^{-1}$. Consequently, $T = (I - W)^{-1} - I$ captures the total causal effects between different variables, aggregating both \textit{direct} and \textit{indirect} effects. The inverse operation $(I - W)^{-1}$ is crucial because it accounts for all possible (direct and indirect) paths through which one variable can affect another. This captures the cumulative effect of all these paths, providing a complete picture of how changes in one variable propagate through the system. See Appendix \ref{sec:epsilon} for further analysis of the properties of $T$. Note that $T $ is only applicable to \textit{linear} causal models, while nonlinear causal models are more complicated (\textcolor{blue}{\cite{pearl-causality-2009}}). 
\end{proof}

To facilitate the explanation of the causal effect matrix $T$, we provide an illustrative example for $T$. Consider a causal model with three variables $X_1$, $X_2$, and $X_3$, where $X_1$ influences $X_2$, and $X_2$ influences $X_3$. The matrix $W$ is represented as follows:
\[
W = \begin{pmatrix}
0 & w_{12} & 0 \\
0 & 0 & w_{23} \\
0 & 0 & 0
\end{pmatrix}.
\]
Here, $w_{12}$ is the direct causal effect of $X_1$ on $X_2$, and $w_{23}$ is the direct causal effect of $X_2$ on $X_3$. The total effect matrix $T$ would include not just these direct causal effects but also the indirect causal effect of $X_1$ on $X_3$ through $X_2$. Visually, this could be represented as:
\begin{center}
$ X_1 \to X_2 \to X_3 $.
\end{center}
In this case, $T_{13}$ captures the indirect causal effect of $X_1$ on $X_3$ through $X_2$, which is not captured by the matrix $W$ alone. To compute the total causal effect matrix $T$, we follow Equation~\ref{eq:causal_effect_matrix} and proceed step by step: first, we calculate $I - W$:

\[
I - W = 
\begin{pmatrix}
1 & 0 & 0 \\
0 & 1 & 0 \\
0 & 0 & 1
\end{pmatrix} - 
\begin{pmatrix}
0 & w_{12} & 0 \\
0 & 0 & w_{23} \\
0 & 0 & 0
\end{pmatrix} 
= 
\begin{pmatrix}
1 & -w_{12} & 0 \\
0 & 1 & -w_{23} \\
0 & 0 & 1
\end{pmatrix}.
\]

\noindent Next, we compute $(I - W)^{-1}$:
\[
(I - W)^{-1}  = I + W + W^2 = \begin{pmatrix}
1 & w_{12} & w_{12}w_{23} \\
0 & 1 & w_{23} \\
0 & 0 & 1
\end{pmatrix}.
\]

\noindent Finally, we subtract the identity matrix $I$ from $(I - W)^{-1}$ to obtain $T$:

\[
T = \begin{pmatrix}
1 & w_{12} & w_{12}w_{23} \\
0 & 1 & w_{23} \\
0 & 0 & 1
\end{pmatrix} - 
\begin{pmatrix}
1 & 0 & 0 \\
0 & 1 & 0 \\
0 & 0 & 1
\end{pmatrix} 
= \begin{pmatrix}
0 & w_{12} & w_{12}w_{23} \\
0 & 0 & w_{23} \\
0 & 0 & 0
\end{pmatrix}.
\]

\noindent Thus, the matrix $T$ captures both the direct causal effects $w_{12}$ and $w_{23}$, as well as the indirect causal effect of $X_1$ on $X_3$, which is $w_{12}w_{23}$.

\noindent\textbf{Remark:} While interventional constraints are introduced here in the context of linear models, they are conceptually general and can be adapted to nonlinear settings. In such cases, total causal effects would be estimated through path-specific derivatives or interventional distributions, though practical implementation would require further research.

\section{Two-Stage Constrained Optimization}\label{sec:method}
We propose a two-stage optimization strategy to solve the causal discovery problem under both acyclicity and interventional constraints. The optimization problem is highly non-convex due to the interplay between structural and parametric constraints. To address this, we propose a practical two-stage constrained optimization approach that combines L-BFGS-B with Sequential Least Squares Programming (SLSQP).

\subsection{Overview of the Optimization Problem}\label{sec:analysis}
In our problem, the Frobenius norm term $\frac{1}{2n} \|X - XW\|_F^2$ is a quadratic function in $W$, and since the trace of a quadratic form is \textit{convex}, this term is \textit{convex}. The $\ell_1$ norm $\lambda \|W\|_1$ is also \textit{convex}. Therefore, the objective function $F(W)$ is \textit{convex}, as it is a sum of \textit{convex} functions. However, the causal effect constraints $\delta_{ij}\,\bigl(T_{i,j} - \delta_{ij}\bigr) > 0$ involve the inverse $(I - W)^{-1}$, a \textit{non-convex} operation. Therefore, these causal effect constraints are \textit{non-convex}. Additionally, the acyclicity constraint $\text{tr}(e^{W \circ W}) - d = 0$ involves an element-wise exponential function $e^{W \circ W}$, which is \textit{convex}. The condition that the trace of this matrix minus a constant equals zero is a typically \textit{non-convex} equality constraint. As a result, although the objective function $F(W)$ is \textit{convex}, the constraints involving the matrix $T$ and the acyclicity condition introduce \textit{non-convexity}, , making the overall optimization problem defined by Equations (\textcolor{blue}{\ref{eq:obj_fun}–\ref{eq:h_w_expression}}) a \textit{non-convex} problem. Furthermore, there are intrinsic tensions between the acyclicity constraint and the interventional constraints, manifested in three key ways: \textit{First}, negative elements in the weight matrix $W$ does not affect $h(W)$ because the Hadamard product $W \circ W$ involves squaring the elements of $W$, which converts all negative values to positive values. Consequently, $W \circ W$ is always non-negative, ensuring that the matrix exponential $e^{W \circ W}$ and its trace are non-negative. Therefore, the value of $h(W)$ is not directly influenced by whether the elements of $W$ are negative or positive. However, negativity of elements in the weight matrix $W$ can impact the causal effect between variables, thus deciding violation of interventional constraints. \textit{Second}, magnitude of elements in the weight matrix has different impact on acyclicity constraints $h(W)$ and interventional constraints. Acyclicity constraints encourage lower values in the weight matrix, while interventional constraints increase the value of elements in weight matrix. \textit{Third}, acyclic constraints encourage a sparse graph, while interventional constraints promote a less sparse graph, depending on the number of interventional constraints and the magnitude of the relevant thresholds $\delta$. For all these reasons, the overall optimization problem defined by Equations (\textcolor{blue}{\ref{eq:obj_fun}–\ref{eq:h_w_expression}}) is not only \textit{non-convex} but also \textit{highly non-convex}, making standard optimizers such as L-BFGS-B insufficient and unreliable for handling the full set of constraints. Therefore, we adopt the Sequential Least Squares Programming (SLSQP) method (\textcolor{blue}{\cite{SLSQP}}), which supports general nonlinear constraints and provides a practical and effective solution for our setting. Given that the SLSQP method is gradient-based, it is essential to compute the gradients of both the objective function $F(W)$ and the constraints. The gradient of the Frobenius norm squared term is:
\begin{equation}
\nabla_W \left( \frac{1}{2n}\|X - XW\|_F^2 \right) = \frac{1}{n} X^T(XW - X)
\end{equation}
and the gradient of the $L_{1}$ norm is:
\begin{equation}
\nabla_W \|W\|_1 = \text{sign}(W),
\end{equation}
where $\text{sign}(W)$ is applied element-wise. The full gradient of the objective function $F(W)$ is then:
\begin{equation}
\nabla F(W) = \frac{1}{n} X^T(XW - X) + \lambda \text{sign}(W).
\end{equation}
The gradient of the causal effect measure $T$ is:
\begin{equation}
\nabla_W T = -(I - W)^{-1} \otimes (I - W)^{-1}.
\label{eq:delta_t}
\end{equation}
The gradient of the acyclicity measure \( h(W) \) is:
\begin{equation}
\nabla_W h(W) = 2 \cdot \text{diag}(e^{W \circ W}) \cdot (W \circ W) \cdot W.
\label{eq:h_gradient_w}
\end{equation}

\noindent The SLSQP method approximates the problem locally by a quadratic model of the objective function and a linear model of the constraints:
\begin{equation}
\min_{\Delta W} \left( \nabla F(W)^T \Delta W + \frac{1}{2} \Delta W^T H \Delta W \right)
\label{eq:objective_on_w}
\end{equation}
subject to
\begin{equation}
A \Delta W = b - c,
\label{eq:a_delta}
\end{equation}
where $H$ is an approximation to the Hessian of $F(W)$. $A$ represents the Jacobians of the interventional and acyclicity constraints from Equations (\textcolor{blue}{\ref{eq:delta_t}-\ref{eq:h_gradient_w}}). $b - c$ represents the amount by which the current constraint values deviate from their desired target values, helping to define the feasible region. $\Delta W$ is the step direction, representing the change in $W$ that minimizes the objective function (Equation \ref{eq:objective_on_w}) while satisfying the constraints (Equation \ref{eq:a_delta}). Using the step direction $\Delta W$ found from solving the quadratic subproblem defined by Equations (\textcolor{blue}{\ref{eq:objective_on_w}-\ref{eq:a_delta}}), the weights are updated as:
\begin{equation}
W \leftarrow W + \alpha \Delta W,
\end{equation}
where $\alpha$ is the step size determined by a line search. 

The SLSQP algorithm starts with an initial weight matrix $W^{(1)}$ and computes the objective function and Jacobians. In the main loop, it iteratively solves a quadratic subproblem to find the step direction $\Delta W$, updating the weight matrix to minimize the objective function while meeting constraints. After each iteration, the algorithm updates $W$, checks for convergence based on the tolerance $tol$, and stops if the change in $W$ is small enough or if $max\_iter$ is reached. The matrix $W_{est}$ is returned as the output. The detailed procedure of SLSQP optimization is outlined in Algorithm \ref{alg:slsqp}. In this paper, the maximum number of iterations, $max\_iter$, is set to 10,000, and the tolerance, $tol$, is set to $1 \times 10^{-6}$. The bounds on the entries of the weight matrix $\mathcal{B}$ are defined as follows:

\begin{equation}
\mathcal{B} = \left\{
\begin{array}{ll}
(0, 0) & \text{for } i = j, \\
(-\infty, \infty) & \text{for } i \neq j,
\end{array}
\right.
\quad i, j \in \{1, 2, \dots, d\}.
\end{equation}
In other words, the diagonal entries (where $i = j$) are constrained to be 0, while the off-diagonal entries (where $i \neq j$) are unbounded.

Once SLSQP produces an estimated weight matrix $W_{\text{est}}$, entries whose absolute values are below $\omega$ are set to zero, making the matrix sparse. However, the estimated weight matrix that satisfies both acyclicity and interventional constraints before thresholding may still fail to fully meet these constraints after thresholding, particularly the interventional constraints. This occurs because thresholding can make the weight matrix sparse, thereby disconnecting parts of the causal edges. Consequently, thresholding may sever causal paths between cause and target variables or weaken their causal strength, leading to violations of some interventional constraints. To address this, one can increase the thresholds $\delta_{ij}$ in the constrained optimization step for any interventional constraints found to be violated post-thresholding. For instance, if variable $i$ is known to have a positive causal effect on variable $j$, the corresponding constraint is $\delta_{ij}\,\bigl(T_{i,j} - \delta_{ij}\bigr) > 0$ with $\delta_{ij}$ initially set to be a small positive value (e.g., $\delta_{ij}=0.01$). If the constraint $\delta_{ij}\,\bigl(T_{i,j} - \delta_{ij}\bigr) > 0$ is satisfied before thresholding but violated after thresholding, we re-optimize with modified deltas as $\delta_{ij} \leftarrow \delta_{ij} + \epsilon, \epsilon > 0$. See Appendix \ref{sec:epsilon} for details on how to choose $\epsilon$. Note that a larger $\delta_{ij}$ can substantially change the learned model, a larger $\delta_{ij}$ imposes stricter constraints that force the model to retain or strengthen more connections. In high-dimensional settings, interventional constraints are also more likely to be violated by thresholding, since longer and more complex causal paths mean that removing any edge can disrupt global causal paths and causal effects between variables. 

\subsection{Two-stage Constrained Optimization}\label{sec:batch_optimization}

 The SLSQP method is sensitive to the initial guess, specifically the starting weight matrix, \( W^{(1)} \), which is particularly problematic in non-convex spaces. Thus, a robust approach is required to ensure convergence to a feasible solution. To address this, we propose a straightforward two-stage constrained optimization approach:

\textbf{Stage One (Optimization without interventional constraints):} Initially, the efficient gradient-based L-BFGS-B algorithm (\textcolor{blue}{\cite{NOTEARS}}) is used to learn a weight matrix $W^{(1)}$ that satisfies the acyclicity constraint. $W^{(1)}$ serves as an initial approximation for the subsequent continuous optimization that further includes interventional constraints. 

\textbf{Stage Two (Optimization with interventional constraints):} The weight matrix $W_0$ is then used as the initial guess for the SLSQP optimization. In this stage, the objective is to iteratively refine the solution to further satisfy the interventional constraints. These interventional constraints are addressed sequentially, ensuring that the solution converges to a feasible and optimal $W^*$.

Our overall two-stage constrained optimization method, Linear Causal Discovery with Interventional Constraints (Lin-CDIC), is summarized in Algorithm \ref{alg:lin-cdic}.

\begin{algorithm}[htbp]
\caption{Lin-CDIC Algorithm}
\label{alg:lin-cdic}
\begin{algorithmic}[1]
\Require Observational data $X$, cause variable set $\mathcal{C}$, target variable set $\mathcal{T}$, acyclicity tolerance $h_{tol}$, weight threshold $\omega$, adjustment factor $\epsilon$
\Ensure Optimal weight matrix $W^*$
\State $ConSat \gets \text{False}$ \Comment{Satisfaction of interventional constraints}
\State $W^{(1)} \gets \text{L-BFGS-B}(X,\, h_{tol})$
\State $\delta \gets \{\delta_{ij}\mid i \in \mathcal{C}, j \in \mathcal{T}\}$ \Comment{Interventional constraint thresholds}
\State $\mathcal{I} \gets \emptyset$ \Comment{Accumulated interventional constraints}
\For{each $i \in \mathcal{C}$ and $j \in \mathcal{T}$}
    \State $\mathcal{I} \gets \mathcal{I} \cup \{\delta_{ij}\,\bigl(T_{i,j} - \delta_{ij}\bigr) > 0\}$
    \While{True}
        \State $W_{\text{est}} \gets \text{SLSQP}(F(W),\, X,\, W^{(1)},\, \delta,\, \mathcal{I})$
        \State $W^* \gets W_{\text{est}} \circ \mathbf{1}(|W_{\text{est}}| > \omega)$
        \State $ConSat \gets \text{Constraint\_check}(W^*,\, \mathcal{I})$
        \If{$W^*$ is a DAG}
            \If{$ConSat$ is \text{True}}
                \State $W^{(1)} \gets W_{\text{est}}$
                \State \textbf{break}
            \Else
                \State $\delta_{ij} \gets \delta_{ij} + \epsilon$ \Comment{Interventional constraint threshold adjustment}
            \EndIf
        \Else
            \State $h_{tol} \gets h_{tol} \times 0.25$
        \EndIf
    \EndWhile
\EndFor
\State \Return $W^*$
\end{algorithmic}
\end{algorithm}

\subsection{Convergence analysis}\label{sec:convergence_analysis}
\begin{tcolorbox}[colback=white, colframe=black, title=\textbf{Proposition 4.1} (Convergence of the Two-Stage Optimization)]
The solution $W^*$ obtained by the two-stage optimization method is a KKT (Karush-Kuhn-Tucker) point of the problem defined by Equations (\textcolor{blue}{\ref{eq:obj_fun}}--\textcolor{blue}{\ref{eq:T_h_F}}).
\end{tcolorbox}

\begin{proof}: In Stage One, since $F$ is twice continuously differentiable, L-BFGS-B converges to a stationary point, satisfying

\begin{gather}
\nabla F(W^{(1)}) + \rho \nabla h(W^{(1)}) = 0.
\label{eq:proof_stage_one}
\end{gather}
However, $W^{(1)}$ may satisfy the acyclicity constraint but not the interventional constraints. In Stage Two, using $W^{(1)}$ as the initialization, SLSQP, by sequential quadratic programming, iteratively updates $W$, producing a sequence $W^{(k)} \to W^*$. As $F$, $h$, and $T_{ij}$ are continuously differentiable and the constraint qualification holds in the feasible region, by the theory of constrained optimization (\textcolor{blue}{\cite{NocedalWright2006}}), the limit point $W^*$ satisfies the following KKT (Karush-Kuhn-Tucker) conditions. Specifically, there exist Lagrange multipliers $\mu \in \mathbb{R}$, $\lambda_{ij} \geq 0$ such that

\begin{gather}
\nabla F(W^*) + \mu^T \nabla h(W^*) + \sum_{(i, j)} \lambda_{ij} \delta_{ij} \nabla T_{ij}(W^*) = 0, \\
h(W^*) = 0, \\
\delta_{ij}(T_{ij}(W^*) - \delta_{ij}) > 0, \\
\lambda_{ij} \cdot [\delta_{ij}(T_{ij}(W^*) - \delta_{ij})] = 0, \quad \forall (i, j).
\end{gather}

\noindent Therefore, the solution $W^*$ obtained by the two-stage optimization method is a KKT point of the original constrained problem (but is not necessarily a global optimum).
\end{proof}

The two-stage approach progressively refines the solution by breaking the optimization process into manageable steps. In the first stage, an initial feasible solution $W^{(1)}$ is obtained that satisfies acyclicity constraint, providing a solid foundation for further refinement, even though it does not yet meet all constraints. This ensures that subsequent optimizations focus on fine-tuning rather than large-scale corrections. In the second stage, the solution is incrementally improved, moving towards the optimal weight matrix $W^{*}$ that satisfies both the acyclicity and interventional constraints. This step-by-step refinement preserves feasibility while progressively approaching the optimal solution.

\subsection{Time Complexity}\label{sec:time_complexity}

The Lin-CDIC method involves two sequential optimization stages: first, an L-BFGS-B gradient-based method, and then SLSQP. The overall computational complexity depends on the number of nodes $d$, the number of interventional constraints $m$, and the nature of the optimization algorithms used. In the first stage, the time complexity is primarily driven by the number of nodes $d$ and the complexity of the underlying gradient-based optimization, which is generally $O(d^3)$ due to the matrix operations involved in enforcing the acyclicity constraint. In the second stage, since each constraint is addressed sequentially, the complexity is linear with respect to the number of interventional constraints, denoted as $m$. Thus, the overall time complexity for this stage can be approximated as $O(m \cdot T_{\text{SLSQP}})$, where $T_{\text{SLSQP}}$ is the time complexity of a single SLSQP iteration, which itself depends on the problem size $d$ and can range from $O(d^2)$ to $O(d^3)$. Combining both stages, the overall time complexity of the batch-constrained optimization method is $O(d^3) + O(m \cdot T_{\text{SLSQP}})$, upper bounded by $(m+1)O(d^3)$. Since $m$ can be large in practical applications, the method's time complexity is effectively linear with respect to $m$.

\subsection{An Illustrative Example for the Problem and Algorithm} \label{sec:Illustrative_problem}
\begin{figure*}[ht]
  \centering
  \begin{minipage}{0.32\linewidth}
    \centering
    \includegraphics[width=\linewidth]{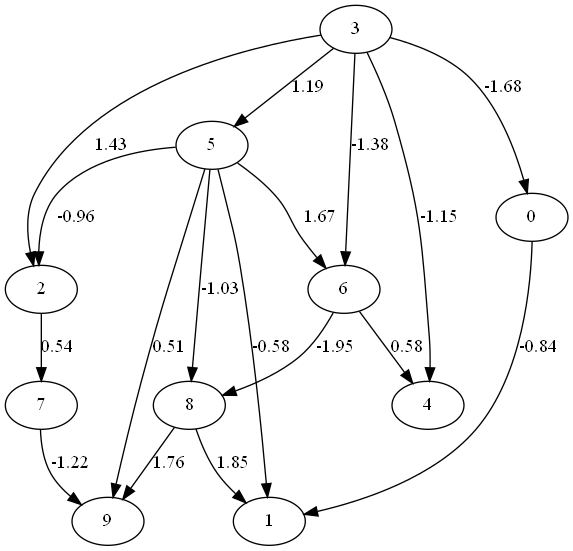}
    \\(a)
    \label{fig:exp1}
  \end{minipage}
  \hfill
  \begin{minipage}{0.32\linewidth}
    \centering
    \includegraphics[width=\linewidth]{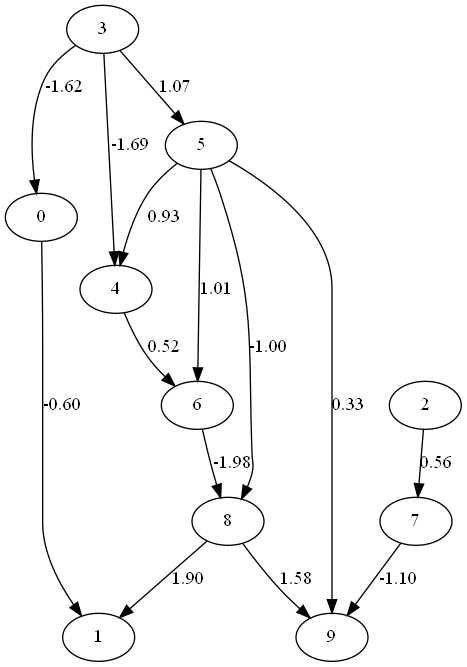}
    \\(b)
    \label{fig:exp2}
  \end{minipage}
  \hfill
  \begin{minipage}{0.32\linewidth}
    \centering
    \includegraphics[width=\linewidth]{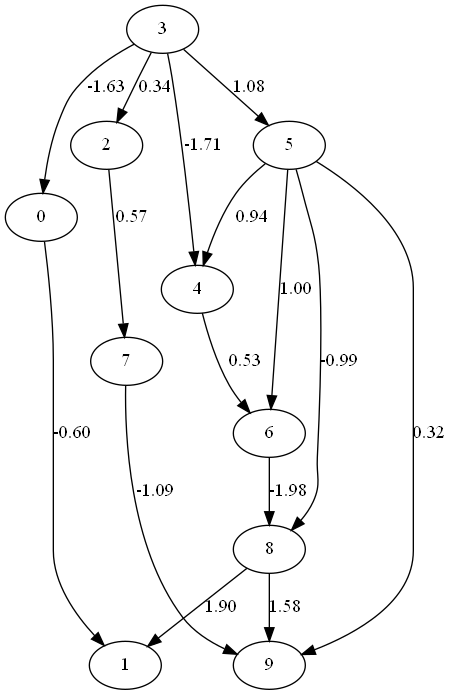}
    \\(c)
    \label{fig:exp3}
  \end{minipage}
  \caption{From left to right: (a) True causal model, (b) Causal model learned without interventional constraints, (c) Causal model learned with interventional constraints.}
  \label{fig:CDIC_illustration}
\end{figure*}

To illustrate the difference between models learned with and without interventional constraints, we provide an example of a linear causal model with 10 variables. We generated data with a sample size of 10 and four interventional constraints: $T(8,9)>0$, $T(3,7)>0$, $T(3,2)>0$, and $T(2,7)>0$, based on the true causal model. Note that we chose a small sample size of 100 specifically to highlight the benefit of incorporating constraints, which is a common practice in studies that consider prior knowledge. The true causal model and the learned models without interventional constraints (i.e., after Stage One) and with interventional constraints (i.e., after Stage Two), are shown in Figure \textcolor{blue}{\ref{fig:CDIC_illustration}}, and the performance metrics (see Section \textcolor{blue}{\ref{sec:Baseline_and_Metrics}} for details) of the learned models are summarized in Table \textcolor{blue}{\ref{tab:illustration_performance}} (better metrics are shown in bold and blue).

\begin{table}[htbp]
\centering
\small
\begin{tabular}{|c|c|c|}
\hline
Metric & Without Interventional Constraints & With Interventional Constraints \\ \hline
\text{FDR} & 0.143 & \textbf{\textcolor{blue}{0.133}} \\ \hline
\text{TPR} & 0.706 & \textbf{\textcolor{blue}{0.765}} \\ \hline
\text{FPR} & \textbf{\textcolor{blue}{0.071}} & \textbf{\textcolor{blue}{0.071}} \\ \hline
\text{SHD} & 6 & \textbf{\textcolor{blue}{5}} \\ \hline
\text{SID} & 9 & \textbf{\textcolor{blue}{7}} \\ \hline
\text{NNZ} & 14 & \textbf{\textcolor{blue}{15}} \\ 
\hline
\text{Time} & \textbf{\textcolor{blue}{3.01}} & 14.59 \\ \hline
\end{tabular}
\caption{Performance metrics of the causal models learned with and without interventional constraints.}
\label{tab:illustration_performance}
\end{table}

In the causal model learned by NOTEARS without interventional constraints (i.e., from Stage One), we observe $T(8,9)=1.578$, $T(3,7)=0$, $T(3,2)=0$, and $T(2,7)=0.563$. As the causal effects from $X_{3}$ to $X_{7}$ and from $X_{3}$ to $X_{2}$ are zero, the conditions $T(3,7)>0$ and $T(3,2)>0$ are violated. In contrast, the model learned with interventional constraints (i.e., from Stage Two) yields $T(8,9)=1.583$, $T(3,7)=0.195$, $T(3,2)=0.343$, and $T(2,7)=0.570$, satisfying all required conditions. Notably, incorporating interventional constraints: (a) correctly recovered the causal paths from $X_{3}$ to $X_{7}$ and $X_{3}$ to $X_{2}$, and (b) adjusted their causal effects from zero to positive. These results demonstrate that interventional constraints influence both the structural and parametric aspects of causal discovery.

\section{Experiments}\label{sec:experiments}

\subsection{Performance Metrics and Baseline Methods}\label{sec:Baseline_and_Metrics}
We conducted experiments on both synthetic and real-world datasets\footnote{The reproducible code and datasets are available at \url{https://github.com/ZhigaoGuo/Lin-CDIC}.}. All experiments were conducted on a laptop running \textit{Windows 11 Home} (version 22H2, build 22631), equipped with a 13th Gen Intel\textsuperscript{\textregistered} Core\texttrademark~i9-13900H processor (14 cores, 20 threads, 2.6\,GHz), 32\,GB of RAM, and a 1\,TB SSD. To evaluate the learned causal models, we consider metrics including False Discovery Rate (FDR), True Positive Rate (TPR), False Positive Rate (FPR), Structural Hamming Distance (SHD) (\textcolor{blue}{\cite{SHD}}), Structural Intervention Distance (SID) (\textcolor{blue}{\cite{SID}}), the Number of Non-Zero entries (NNZ), i.e. number of causal edges, and time (in seconds). For the above metrics, lower is better, except for TPR, for which higher is better. In addition to the previously introduced metrics, we assess the estimated matrix by comparing the signs of its elements with those of the true weight matrix. This measure is referred to as the Sign Consistency Sum (SCS). Specifically, let $W_{\text{est}}$ and $W_{\text{true}}$ be the estimated and true weight matrices, both of dimension $d \times d$. The Sign Consistency Sum is defined as:
\begin{equation}
\text{SCS}(W_{\text{est}}, W_{\text{true}}) = \sum_{i=1}^d \sum_{j=1}^d \mathbf{1}_{\{\operatorname{sgn}(W_{\text{est}, ij}) = \operatorname{sgn}(W_{\text{true}, ij})\}},
\end{equation}
where $\operatorname{sgn}(x)$ is the sign function, defined as $\operatorname{sgn}(x) = 1$ if $x > 0$, $\operatorname{sgn}(x) = 0$ if $x = 0$, and $\operatorname{sgn}(x) = -1$ if $x < 0$. SCS ranges from 0 to $d^2$, which is the number of elements in $W_{\text{true}}$ or $W_{\text{est}}$. A high SCS indicates that the positive and negative influences between variables are accurately captured, preserving the nature of causality — whether one variable \textit{increases} (or \textit{decreases}) as a result of another. This is particularly important in domains such as gene regulatory networks, where the sign of causal influence (\textit{activation} or \textit{inhibition}) can determine the behavior of complex biological systems. Thus, a high SCS enhances the trustworthiness of the model in practical applications, making it a critical metric for assessing the quality of causal inferences. Since no existing method supports the newly introduced interventional constraints, we demonstrate their value by comparing causal models learned with and without these constraints. We also compare with causal models learned with structural path constraints. For continuous optimization-based causal discovery, path constraints can represented using the reachability matrix, 
\begin{equation}
R = \left(I + \frac{\tanh(W)}{d}\right)^d,
\label{eq:reachability_matrix}
\end{equation} 
where $d$ denotes the number of variables. When $d=1$, $R_{ij}>0$ indicates direct reachability between variable pairs $i$ and $j$, i.e., edge constraints. In contrast, when $d>1$, $R_{ij}>0$ indicates indirect reachability between variable pairs $i$ and $j$, i.e., path constraints. See Appendix \ref{sec:cd_path_algorithm} for further analysis of the properties of $R$. To illustrate the difference between path constraints measured by $R$ and interventional constraints by $T$, consider the case where variable $i$ has a \textit{negative} causal effect on variable $j$, the corresponding interventional constraint is given by $T_{ij}<0$, while the associated path constraint is $R_{ij}>0$. Linear Causal Discovery with Path constraints (Lin-CD-Path) is optimized using our two-stage procedure, except that the metric $T_{ij}, i \in \mathcal{C}, j \in \mathcal{T}$ is replaced with $R_{ij}, i \in \mathcal{C}, j \in \mathcal{T}$. The details of the Lin-CD-Path algorithm are summarized in Algorithm \ref{alg:cd_path} in Appendix \ref{sec:cd_path_algorithm}). Thus, in summary, we compare the performance of three methods: (A) \textbf{NOTEARS} that does not incorporate any constraints, including path or interventional constraints; (B) \textbf{Lin-CD-Path} that incorporates causal path constraints; and (C) \textbf{Lin-CDIC} method that incorporates interventional constraints. By contrasting the learned models from (A), (B), and (C), we aim to highlight the unique benefits of incorporating interventional constraints into causal discovery. For all methods, the threshold is set to $\omega=0.3$, consistent with other continuous optimization approaches (\textcolor{blue}{\cite{NOTEARS}}).

\subsection{Synthetic Experiments} \label{sec:syn_exp}
We generate random \textit{linear} causal models characterized by scale-free (SF) graphs (\textcolor{blue}{\cite{Scale-free}}) with Gaussian noise. The number of causal edges is randomly selected between eight and $\min\left(\left\lfloor \frac{d(d - 1)}{2} \right\rfloor, 10\right)$, where $d$ denotes the number of nodes. As for the interventional constraints, we sample from the true causal model based on the strength of the causal effects between cause and target variables. A causal effect from variable $i$ to $j$, denoted as $T_{ij}$, is considered significant if $|T_{ij}| > 0.1$ and is likely to be sampled. The above definition has real-world implications in fields such as genomics, econometrics, and systems biology. For example, weak causal effects are often seen as potentially spurious connections. 

\subsubsection{Effect of Sample Size under Fixed Constraints} \label{sec:syn_sample_size}
\noindent\textbf{Setting}: Firstly, to explore the impact of varying data sizes on constraint satisfaction, we conduct experiments under a fixed number of interventional constraints. In these experiments with 20 variables, the number of constraints was set to two, and the data sizes were varied as 50, 100, 150, and 200. For each setting, we ran 20 experiments. The performance of two methods is shown in Table \ref{tab:details_across_sample_size}. Better metrics are shown in bold and blue. Note that, the sample sizes were deliberately kept small, with $n \in \{50, 100, 150, 200\}$, motivated by recent research such as \textit{Sample Complexity Bounds for Score-Matching: Causal Discovery and Generative Modeling} (\textcolor{blue}{\cite{Sample_complexity}}). This work provides a theoretical analysis of sample complexity bounds in causal discovery and shows that, for causal models with low nonlinearity (quantified by $C_m$, where $C_m = 0$ corresponds to linear models), the SHD between the learned and true causal models decreases significantly as the sample size increases. Intuitively, Table 2 in \cite{Sample_complexity} highlights the relationship between sample complexity and model size for causal models with $C_m = 1$ and 10 variables. This setting corresponds to causal models that are nearly linear, showing that the mean SHD drops from 32 to 13 as the sample size increases from 5 to 160. These insights, derived from simulations of causal discovery without interventional constraints, justify our use of low sample sizes to evaluate the effectiveness of our proposed method.

\begin{table}[ht]
\centering
\footnotesize
\setlength{\tabcolsep}{3pt}
\renewcommand{\arraystretch}{0.85}
\begin{tabular}{@{}cccccc@{}}
\toprule
Methods & Metrics & $n=50$ & $n=100$ & $n=150$ & $n=200$ \\ 
\midrule
\multirow{9}{*}{\shortstack{NOTEARS \\ (Without \\ Constraints)}} 
    & FDR  & (0.113, 0.013) & (0.025, 0.002) & (0.037, 0.004) & (0.030, 0.002) \\
    & TPR  & (0.892, 0.005) & (0.883, 0.004) & (0.898, 0.001) & (0.886, 0.002) \\
    & FPR  & (0.009, 0.000) & (0.002, 0.000) & (0.002, 0.000) & (0.002, 0.000) \\
    & SHD  & (2.700, 4.410) & (1.650, 0.728) & (1.350, 0.328) & (1.600, 0.940) \\
    & SID  & (6.200, 44.460) & (3.350, 5.528) & (3.550, 2.848) & (3.450, 8.648) \\
    & SCS  & 7,939 & 7,965 & 7,966 & 7,961 \\
    & NNZ  & (\textbf{\textcolor{blue}{13.700, 14.910}}) & (12.500, 14.550) & (12.100, 6.590) & (12.700, 10.710) \\
    & Time & \textbf{\textcolor{blue}{6.2}} & \textbf{\textcolor{blue}{4.1}} & \textbf{\textcolor{blue}{2.7}} & \textbf{\textcolor{blue}{3.5}} \\
\midrule
\multirow{9}{*}{\shortstack{Lin-CD-Path \\ (With \\ Path \\ Constraints)}} 
    & FDR  & (0.124, 0.018) & (0.046, 0.005) & (0.049, 0.003) & (0.045, 0.003) \\
    & TPR  & (0.937, 0.007) & (0.947, 0.006) & (0.948, 0.002) & (0.939, 0.003) \\
    & FPR  & (0.011, 0.000) & (0.003, 0.000) & (0.003, 0.000) & (0.003, 0.000) \\
    & SHD  & (2.450, 7.048) & (1.150, 1.928) & (1.150, 1.028) & (1.150, 1.628) \\
    & SID  & (4.450, 32.348) & (1.250, 3.888) & (1.450, 2.848) & (2.200, 7.660) \\
    & SCS  & 7,944 & 7,976 & 7,975 & 7,971 \\
    & NNZ  & (14.550, 15.448) & (\textbf{\textcolor{blue}{13.550, 13.448}}) & (\textbf{\textcolor{blue}{12.950, 7.548}}) & (\textbf{\textcolor{blue}{13.600, 10.440}}) \\
    & Time & 220.3 & 209.9 & 352.4 & 276.2 \\
\midrule
\multirow{9}{*}{\shortstack{Lin-CDIC \\ (With \\ Interventional \\ Constraints)}} 
    & FDR  & (\textbf{\textcolor{blue}{0.094, 0.013}}) & (\textbf{\textcolor{blue}{0.032, 0.004}}) & (\textbf{\textcolor{blue}{0.016, 0.002}}) & (\textbf{\textcolor{blue}{0.021, 0.001}}) \\
    & TPR  & (\textbf{\textcolor{blue}{0.959, 0.007}}) & (\textbf{\textcolor{blue}{0.959, 0.004}}) & (\textbf{\textcolor{blue}{0.971, 0.002}}) & (\textbf{\textcolor{blue}{0.957, 0.003}}) \\
    & FPR  & (\textbf{\textcolor{blue}{0.008, 0.000}}) & (\textbf{\textcolor{blue}{0.002, 0.000}}) & (\textbf{\textcolor{blue}{0.001, 0.000}}) & (\textbf{\textcolor{blue}{0.001, 0.000}}) \\
    & SHD  & (\textbf{\textcolor{blue}{1.800, 5.360}}) & (\textbf{\textcolor{blue}{0.850, 1.428}}) & (\textbf{\textcolor{blue}{0.550, 0.748}}) & (\textbf{\textcolor{blue}{0.750, 1.088}}) \\
    & SID  & (\textbf{\textcolor{blue}{2.900, 11.890}}) & (\textbf{\textcolor{blue}{0.850, 1.528}}) & (\textbf{\textcolor{blue}{0.800, 2.060}}) & (\textbf{\textcolor{blue}{1.450, 4.748}}) \\
    & SCS  & \textbf{\textcolor{blue}{7,960}} & \textbf{\textcolor{blue}{7,982}} & \textbf{\textcolor{blue}{7,988}} & \textbf{\textcolor{blue}{7,981}} \\
    & NNZ  & (14.350, 14.628) & (13.550, 16.050) & (12.800, 7.460) & (13.500, 9.750) \\
    & Time & 622.3 & 300.6 & 553.5 & 425.0 \\
\bottomrule
\end{tabular}
\caption{Performance Metrics Across Sample Sizes (Mean ± Variance). The mean and variance of the edge numbers in the generated causal models, i.e. NNZ, for the four settings are (13.55, 15.25), (13.05, 12.25), (13.20, 8.66), and (13.85, 11.03), respectively.}
\label{tab:details_across_sample_size}
\end{table}

\noindent\textbf{Analysis:} From Table~\ref{tab:details_across_sample_size}, we observe a general trend across all methods: as the sample size increases (with the number of constraints remaining fixed), FDR, FPR, SHD, and SID tend to decrease, while TPR and SCS increase. This indicates the benefit of larger sample sizes for improving causal discovery performance. \textbf{Lin-CDIC} consistently achieves superior results across all metrics. Notably, its SID values—which evaluate the model from a downstream causal inference perspective—are significantly lower than those of the baselines, highlighting the advantages of incorporating interventional constraints. Furthermore, the SCS metric of \textbf{Lin-CDIC}, which reflects the number of correctly recovered signs of causal effects between variables, is higher than that of the baselines, even when only two interventional constraints are used. In contrast, \textbf{NOTEARS} exhibits higher FDR and SHD, particularly when the sample size is small (e.g., $n = 50$), and while \textbf{Lin-CD-Path} provides moderate improvements, it does not match the performance of \textbf{Lin-CDIC}. This may be due to the fact that path constraints are generally less informative than interventional constraints for recovering causal models. In terms of time consumption, \textbf{NOTEARS} is significantly more efficient than both \textbf{Lin-CD-Path} and \textbf{Lin-CDIC}, as it is implemented using efficient L-BFGS-B, which only enforces acyclicity constraints. In contrast, \textbf{Lin-CD-Path} and \textbf{Lin-CDIC} employ more complex SLSQP optimization to handle additional path and interventional constraints. Since path constraints are generally less restrictive than interventional constraints, \textbf{Lin-CD-Path} is consequently more efficient than \textbf{Lin-CDIC}. Note that, since the number of constraints is fixed and the sample size only varies between 50 and 200, the time consumption of each method remains relatively stable, as expected.

\noindent\textbf{Remark:} For experiments with 20 variables, the number of elements in $W_{\text{true}}$ or $W_{\text{est}}$ is 400. Therefore, the maximum possible SCS value across 20 experiments is 8,000. Since the differences after averaging are relatively small, we report the total SCS summed over all 20 experiments. As shown, with two interventional constraints, the causal models learned by \textbf{Lin-CDIC} achieve approximately 20 more correctly signed causal effects than those learned by \textbf{NOTEARS}, and about 10 more than those learned by \textbf{Lin-CD-Path}. This highlights the benefit of incorporating interventional constraints, which contribute not only to structural regularization but also to parameter refinement.

\subsubsection{Effect of Constraints under Fixed Sample Size} \label{sec:syn_constraint_size}
\noindent\textbf{Setting}: To further demonstrate the impact of increasing the number of interventional constraints, we conducted experiments with a fixed amount of data while varying the number of interventional constraints. Specifically, we tested models with 20 variables and a sample size of 100, varying the number of interventional constraints from one to four. Note that the sample size was set to 100 to highlight the benefit of incorporating constraints. The number of constraints was limited to four, as, on one hand, eliciting a large number of constraints is often impractical, and on the other hand, our Lin-CDIC method becomes significantly more time-consuming as the number of constraints increases. For each setting, we ran 20 experiments. The results are shown in Table \ref{tab:details_across_constraint_size}. Better metrics are shown in bold. Note that for each constraint size setting, the generated causal models differ, as increasing the number of constraints may invalidate models that satisfied fewer constraints at lower settings.

\begin{table}[ht]
\centering
\footnotesize
\setlength{\tabcolsep}{3pt}
\renewcommand{\arraystretch}{0.85}
\begin{tabular}{@{}cccccc@{}}
\toprule
Methods & Metrics & $m=1$ & $m=2$ & $m=3$ & $m=4$ \\ 
\midrule
\multirow{9}{*}{\scriptsize \shortstack{NOTEARS \\ (Without \\ Constraints)}} 
    & FDR  & (\textbf{\textcolor{blue}{0.028, 0.003}}) & (0.025, 0.002) & (0.038, 0.003) & (0.037, 0.003)\\
    & TPR  & (0.869, 0.007) & (0.883, 0.004) & (0.880, 0.004) & (0.878, 0.004)\\
    & FPR  & (\textbf{\textcolor{blue}{0.002, 0.000}}) & (0.002, 0.000) & (0.003, 0.000) & (0.003, 0.000)\\
    & SHD  & (1.550, 0.747) & (1.650, 0.728) & (1.950, 2.048) & (1.800, 1.760)\\
    & SID  & (3.800, 20.160) & (3.350, 5.528) & (3.850, 16.428) & (4.100, 14.590)\\
    & SCR  & 7,965 & 7,965 & 7,957 & 7,959\\
    & NNZ  & (11.350, 15.928) & (12.500, 14.550) & (12.750, 16.488) & (12.400, 18.040) \\
    & Time & \textbf{\textcolor{blue}{4.9}} & \textbf{\textcolor{blue}{4.1}} & \textbf{\textcolor{blue}{4.7}} & \textbf{\textcolor{blue}{6.2}}\\
\midrule
\multirow{9}{*}{\scriptsize \shortstack{Lin-CD-Path \\ (With \\ Path \\ Constraints)}} 
    & FDR  & (0.069, 0.007) & (0.046, 0.005) & (0.019, 0.001) & (0.041,  0.002)\\
    & TPR  & (0.934, 0.008) & (0.947, 0.006) & (0.959, 0.004) & (0.937, 0.004)\\
    & FPR  & (0.005, 0.000) & (0.003, 0.000) & (0.002, 0.000) & (0.003, 0.000)\\
    & SHD  & (1.350, 2.428) & (1.150, 1.928) &  (0.950, 2.048) & (1.450, 2.348)\\
    & SID  & (2.300, 18.910) & (1.250, 3.888) & (1.150, 4.728) & (1.550, 5.648)\\
    & SCS  & 7,969 & 7,976 & 7,981 & 7,971\\
    & NNZ  & (12.550, 14.648) & (\textbf{\textcolor{blue}{13.550, 13.448}}) & 13.450, 14.048) & (\textbf{\textcolor{blue}{13.100, 15.490}})\\
    & Time & 144.4 & 209.9 & 276.6 & 406.3\\
\midrule
\multirow{9}{*}{\scriptsize \shortstack{Lin-CDIC \\ (With \\ Interventional \\ Constraints)}} 
    & FDR  & (0.036, 0.006) & (\textbf{\textcolor{blue}{0.032, 0.004}}) & (\textbf{\textcolor{blue}{0.014, 0.001}}) & (\textbf{\textcolor{blue}{0.012, 0.001}})\\
    & TPR  & (\textbf{\textcolor{blue}{0.956, 0.005}}) & (\textbf{\textcolor{blue}{0.959, 0.004}}) & (\textbf{\textcolor{blue}{0.966, 0.003}}) & (\textbf{\textcolor{blue}{0.959, 0.003}})\\
    & FPR  & (0.003, 0.000) & (\textbf{\textcolor{blue}{0.002, 0.000}}) & (\textbf{\textcolor{blue}{0.001, 0.000}}) & (\textbf{\textcolor{blue}{0.001, 0.000}})\\
    & SHD  & (\textbf{\textcolor{blue}{0.850, 1.528}}) & (\textbf{\textcolor{blue}{0.850, 1.428}}) & (\textbf{\textcolor{blue}{0.700, 1.210}}) & (\textbf{\textcolor{blue}{0.700, 1.010}})\\
    & SID  & (\textbf{\textcolor{blue}{1.800, 17.460}}) & (\textbf{\textcolor{blue}{0.850, 1.528}}) & (\textbf{\textcolor{blue}{0.700, 1.510}}) & (\textbf{\textcolor{blue}{0.750, 1.488}})\\
    & SCS  & \textbf{\textcolor{blue}{7,980}} & \textbf{\textcolor{blue}{7,982}} & \textbf{\textcolor{blue}{7,986}} & \textbf{\textcolor{blue}{7,986}}\\
    & NNZ  & (\textbf{\textcolor{blue}{12.500, 15.750}}) & (13.550, 16.050) & (\textbf{\textcolor{blue}{13.500, 14.650}}) & (13.050, 16.348) \\
    & Time & 263.2 & 300.6 & 351.8 & 506.9\\
\bottomrule
\end{tabular}
\caption{Performance Metrics Across Constraint Sizes (Mean ± Variance). The mean and variance of the edge numbers in the generated causal models, i.e. NNZ, for the four settings are (12.50, 13.75), (13.05, 12.25), (13.60, 16.74), and (13.20, 15.46), respectively.}
\label{tab:details_across_constraint_size}
\end{table}

\noindent\textbf{Analysis:} From Table \ref{tab:details_across_constraint_size}, we can conclude that \textbf{Lin-CDIC} consistently achieves the best overall accuracy across nearly all constraint sizes, except when only a single constraint is applied—where the constraining effect is minimal. It achieves the lowest SHD and SID, along with the highest TPR and SCS in each setting, indicating superior recovery of the true causal model. In terms of time consumption, \textbf{NOTEARS} is significantly more efficient than both \textbf{Lin-CD-Path} and \textbf{Lin-CDIC}. Moreover, while \textbf{NOTEARS} remains largely unaffected by the number of constraints, both \textbf{Lin-CD-Path} and \textbf{Lin-CDIC} exhibit a clear increase in runtime as the number of constraints grows. This observation is consistent with the theoretical time complexity analysis presented in Section~\textcolor{blue}{\ref{sec:time_complexity}}, which suggests that \textbf{Lin-CD-Path} and \textbf{Lin-CDIC} become more computationally expensive when more constraints are incorporated.

\noindent\textbf{Remark}: The constrained problem presented in this paper, includes both nonlinear equality constraints that enforce DAG-ness and nonlinear inequality or bound constraints that restrict reachability and the negativity of causal effects between variables. Optimizing such a problem with many constraints is particularly challenging. In our experiments, we observed that standard optimization methods, such as L-BFGS-B, are inadequate, leading us to adopt Sequential Least Squares Programming (SLSQP), which can handle general constraints. As the defined optimization problem is non-convex (see analysis in Section \ref{sec:analysis}), solving it is computationally demanding (see time complexity in Section \ref{sec:time_complexity}). Moreover, since the problem is non-convex, there is no guarantee of finding the globally optimal solution. Consequently, the scalability of our method is limited. Through this work, we aim to inspire further efforts to address the scalability challenges associated with our method. For instance, developing new optimization techniques specifically tailored to interventional constraints could significantly enhance both the scalability and efficiency of our approach.

\subsection{Real-world Experiment}\label{sec:real_exp}
\begin{figure}[ht]
    \centering
    \includegraphics[width=0.4\textwidth]{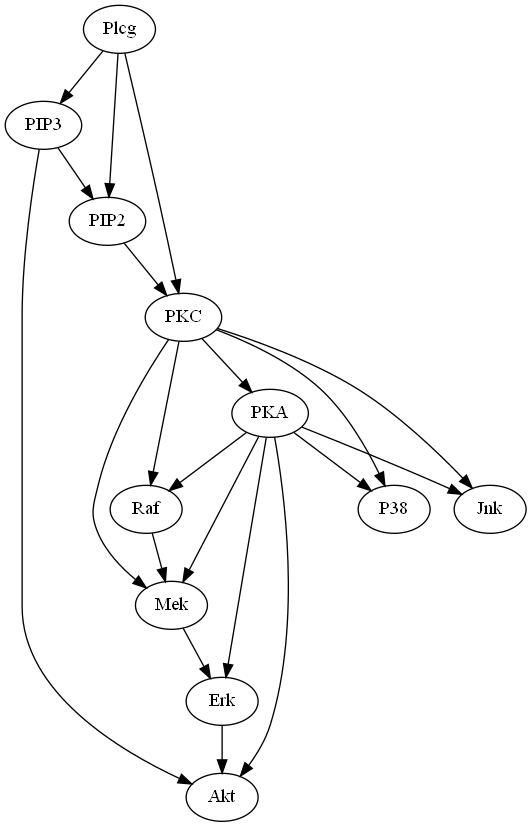}  
    \caption{True Sachs causal graph.}
    \label{fig:true_sachs_graph}
\end{figure}

In addition to synthetic experiments, we also test on the widely used Sachs dataset (\textcolor{blue}{\cite{Sachs-2005}}), which contains both observational and experimental flow cytometry data on protein signaling in human immune cells. Although this is a single dataset, it remains one of the most comprehensive benchmarks for evaluating causal discovery methods. We employ the Sachs causal graph, shown in Figure \ref{fig:true_sachs_graph}, and available at \url{https://www.bnlearn.com/research/sachs05/}, which contains 20 causal edges, as a benchmark, despite controversies arising from uncertainties in intervention specificity, potential cyclic dependencies in cellular signaling networks, unmeasured confounding that challenges causal sufficiency, and discrepancies between the consensus network and the observed experimental data (\textcolor{blue}{\cite{schmidt2009cyclic,Mooij-Sachs-2,Mooij-Sachs-1}}). 

\subsubsection{Sachs Causal Interactions Discussion} \label{sec:prior_dicussion}

As Figure \ref{fig:true_sachs_graph} only indicates causal pathways between proteins without specifying particular causal interactions, such as \textit{inhibition} or \textit{activation}, we augmented the Sachs dataset with causal interactions from the literature and knowledge bases like Reactome (\url{https://reactome.org/}). Among a subset of the 11 phosphorylated proteins and phospholipids, we collected and discussed eight known causal interactions, as detailed below: 

\begin{itemize}

\item \textcolor{blue}{``PKC \textit{activate} JNK''}: In \textcolor{blue}{\citealp{PKC-Jnk-2008}}, the \textit{Abstract} states: ``PKC can augment the degree of JNK activation by phosphorylating JNK...''; the \textit{Results} section notes: ``To achieve a more efficient activation of JNK, phosphorylation by PKC should precede phosphorylation by MKK4 or MKK7.''; and the \textit{Discussion} adds: ``Our data showed that phosphorylation by PKC enhances JNK activation by increasing MKK4/7-dependent phosphorylation.'' Therefore, we can conclude that ``PKC may indirectly \textit{activate} JNK,'' which can be expressed as our interventional constraint: \textcolor{blue}{$T(\text{PKC}, \text{JNK}) > 0$}.

\item \textcolor{blue}{``PKC \textit{activate} P38''}: In \textcolor{blue}{\citealp{PKC-P38-2006}}, the \textit{Results} section notes: ``Thus, it appears that the MEK/ERK and p38 signaling pathways are important downstream effectors of PKC$\delta$ in platelets.'' The \textit{Discussion} section adds: ``We demonstrated that MEK1/2, ERK1/2, and p38 are activated by collagen and thrombin, and more importantly, established the requirement for PKC$\delta$ and PLC activation in this process.'' Finally, the \textit{Conclusion} summarizes: ``PKC$\delta$ then triggers activation of the MEK/ERK and p38 signaling pathways, which ultimately result in the generation and release of TxA$_2$.'' In \textcolor{blue}{\citealp{PKC-P38-2004}}, the \textit{Abstract} also notes: ``PKC$\alpha$ was found to be requisite for the activation of p38MAPK in LPS-stimulated microglia.'' Therefore, we can conclude that ``PKC may indirectly \textit{activate} P38,'' which can be expressed as our interventional constraint: \textcolor{blue}{$T(\text{PKC}, \text{P38}) > 0$}. 

\item \textcolor{blue}{``PIP3 \textit{activates} Akt''}: In \textcolor{blue}{\citealp{PIP3-Akt-2007}}, it is noted that ``PI3K phosphorylates phosphatidylinositol-4,5-bisphosphate (PIP2) to generate phosphatidylinositol-3,4,5-trisphosphate (PIP3), in a reaction that can be reversed by the PIP3 phosphatase PTEN. AKT and PDK1 bind to PIP3 at the plasma membrane, and PDK1 phosphorylates the activation loop of AKT at T308,'' a finding also acknowledged at \url{https://reactome.org/content/detail/R-HSA-1257604} \textcolor{blue}{\citealp{Reactome-2018}}. However, \textcolor{blue}{\citealp{PIP3-Akt-2021}} further suggest that Akt may indirectly inhibit additional PIP3 synthesis through feedback, indicating the presence of a feedback loop between PIP3 and Akt. In our paper, we study causal discovery under the assumption of a Directed Acyclic Graph (DAG), which means that ``PIP3 activates Akt'' and ``Akt inhibits PIP3'' cannot be incorporated simultaneously. Nevertheless, we can at least conclude that ``PIP3 \textit{activates} Akt,'' which can be formalised as our interventional constraint: \textcolor{blue}{$T(\text{PIP3}, \text{Akt}) > 0$}.

\item \textcolor{blue}{``PKA \textit{inhibit} P38''}: In \textcolor{blue}{\citealp{PKA-P38-2021}}, the \textit{Results} section states, ``These results suggest that PKA inhibition in the PA/PDE4/PKA pathway activates p38.'' The \textit{Discussion} further explains, ``We find that decreasing the basal PKA activity through the PA/PDE4/PKA pathway or using direct PKA inhibitors results in p38 and ERK1/2 activation. PKA activity seems then to exert a negative regulation upon p38 and ERK1/2 involved in EGFR endocytosis, which would be released when the PA/PDE4/PKA pathway is stimulated with propranolol.'' Therefore, we can conclude that ``PKA may indirectly \textit{inhibit} P38,'' which can be expressed as our interventional constraint: \textcolor{blue}{$T(\text{PKA}, \text{P38}) < 0$}.

\item \textcolor{blue}{``PKA \textit{inhibit} Raf''}: \textcolor{blue}{\citealp{PKA-Raf-1994}} and \textcolor{blue}{\citealp{PKA-Raf-2003}} consistently report that ``When PKA is activated, it phosphorylates Raf-1 and stimulates recruitment of 14-3-3, preventing Raf-1 recruitment to the plasma membrane and subsequently blocking its activation,'' and ``We also show that endogenous Raf-1 and PKA form a complex that is disrupted when cAMP levels in cells are elevated, and... the PKA inhibitor H89 rescues Raf-1 activation in the presence of forskolin/IBMX.'' In addition, they state that ``PKA can inhibit Raf-1 function directly via phosphorylation of the Raf-1 kinase domain.'' Therefore, we can conclude that ``PKA may directly \textit{inhibit} Raf,'' which can be expressed as our interventional constraint: \textcolor{blue}{$T(\text{PKA}, \text{Raf}) < 0$}.

\item \textcolor{blue}{“Raf \textit{activates} MEK, MEK \textit{activates} ERK, and Raf \textit{activates} ERK,”}: \textcolor{blue}{\citealp{Raf-MEK-2007}} report that “Raf kinases phosphorylate and activate the MEK1 and MEK2 dual-specificity protein kinases,” and “MEK1/2 then phosphorylate and activate the ERK1 and ERK2 MAPKs.” They further note that “Activated ERKs phosphorylate and regulate the activities of an ever-growing roster of substrates...” Based on this cascade, we conclude that “Raf \textit{activates} MEK, MEK \textit{activates} ERK, and thus Raf may indirectly \textit{activate} ERK,” which can be formalised as the following interventional constraints: \textcolor{blue}{$T(\text{Raf}, \text{MEK}) > 0$}, \textcolor{blue}{$T(\text{MEK}, \text{ERK}) > 0$}, and \textcolor{blue}{$T(\text{Raf}, \text{ERK}) > 0$}.

\end{itemize}

The eight causal interactions and their corresponding interventional constraints and path constraints are listed in Table \ref{tab:interventional_constraints}. Note that causal interactions between proteins and phospholipids may be either direct or indirect; our method supports both cases without distinction in the interventional constraints.

\begin{table*}[ht]
\centering
\begin{threeparttable}
\begin{tabular}{|c|c|c|}
\hline
\textbf{Causal Interactions} & \textbf{Interventional Constraints} & \textbf{Path Constraints} \\
\hline
PKC \textit{activates} Jnk & $T(\text{PKC}, \text{Jnk}) > 0$ & $R(\text{PKC}, \text{Jnk}) > 0$ \\
\hline
PKC \textit{activates} P38 & $T(\text{PKC}, \text{P38}) > 0$ & $R(\text{PKC}, \text{P38}) > 0$ \\
\hline
PIP3 \textit{activates} Akt & $T(\text{PIP3}, \text{Akt}) > 0$ & $R(\text{PIP3}, \text{Akt}) > 0$ \\
\hline
PKA \textit{inhibits} P38 & $T(\text{PKA}, \text{P38}) < 0$ & $R(\text{PKA}, \text{P38}) > 0$ \\
\hline
PKA \textit{inhibits} Raf & $T(\text{PKA}, \text{Raf}) < 0$ & $R(\text{PKA}, \text{Raf}) > 0$ \\
\hline
Raf \textit{activates} Erk & $T(\text{Raf}, \text{Erk}) > 0$ & $R(\text{Raf}, \text{Erk}) > 0$ \\
\hline
Raf \textit{activates} Mek & $T(\text{Raf}, \text{Mek}) > 0$ & $R(\text{Raf}, \text{Mek}) > 0$ \\
\hline
Mek \textit{activates} Erk & $T(\text{Mek}, \text{Erk}) > 0$ & $R(\text{Mek}, \text{Erk}) > 0$ \\
\hline
\end{tabular}
\end{threeparttable}
\caption{Causal interactions, interventional constraints, and path constraints in the Sachs dataset.}
\label{tab:interventional_constraints}
\end{table*}

\begin{figure}[h]
    \centering
    \begin{minipage}{0.45\textwidth}
        \centering
        \includegraphics[width=\linewidth]{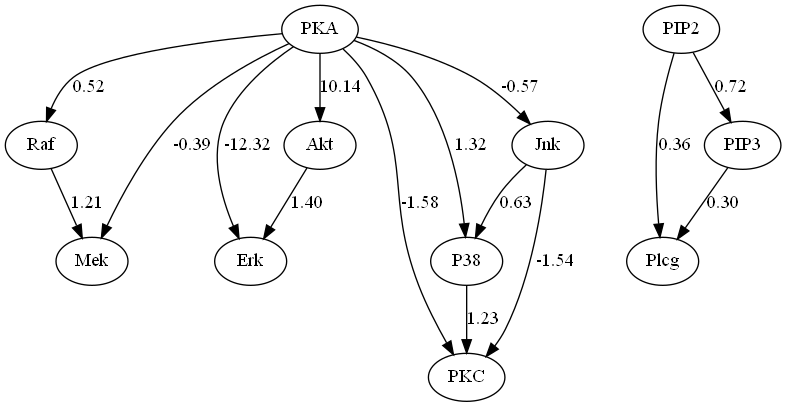}
        \\(a) Without constraints.
    \end{minipage}
    \hfill
    \begin{minipage}{0.45\textwidth}
        \centering
        \includegraphics[width=\linewidth]{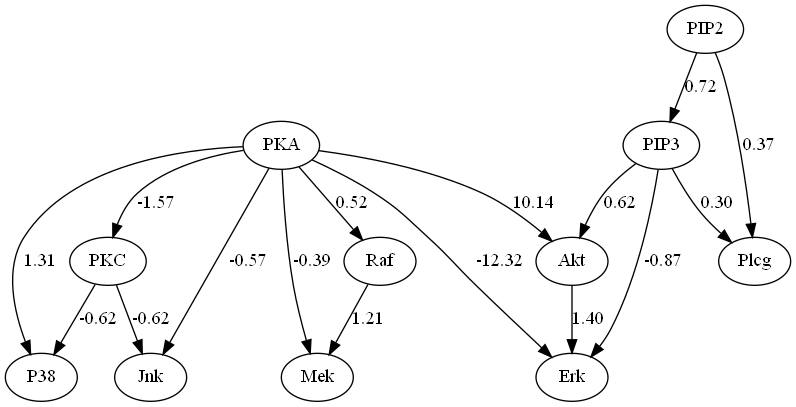}
        \\(b) With path constraints.
    \end{minipage}
    \caption{Sachs causal models learned by NOTEARS (without constraints) and Lin-CD-Path (with path constraints).}
    \label{fig:sachs_model_notears_lin_cd_path}
\end{figure}

\begin{figure}[h]
    \centering
    \begin{minipage}{0.45\textwidth}
        \centering
        \includegraphics[width=\linewidth]{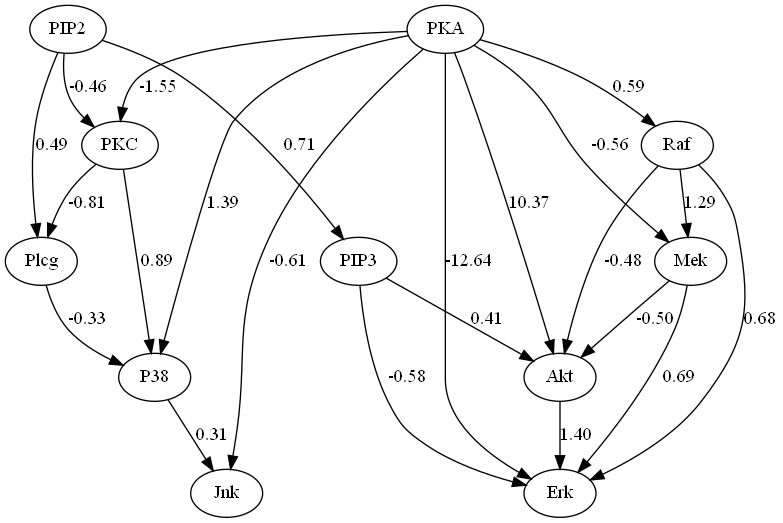}
        \\(a) $\epsilon=0.25$
    \end{minipage}
    \hfill
    \begin{minipage}{0.45\textwidth}
        \centering
        \includegraphics[width=\linewidth]{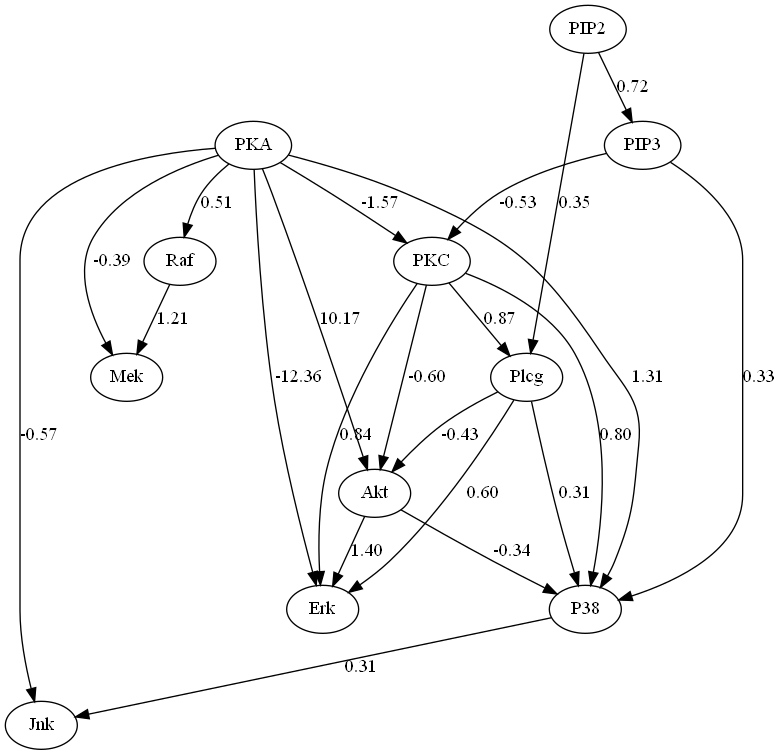}
        \\(b) $\epsilon=0.50$
    \end{minipage}
    
    \vspace{1em}

    \begin{minipage}{0.45\textwidth}
        \centering
        \includegraphics[width=\linewidth]{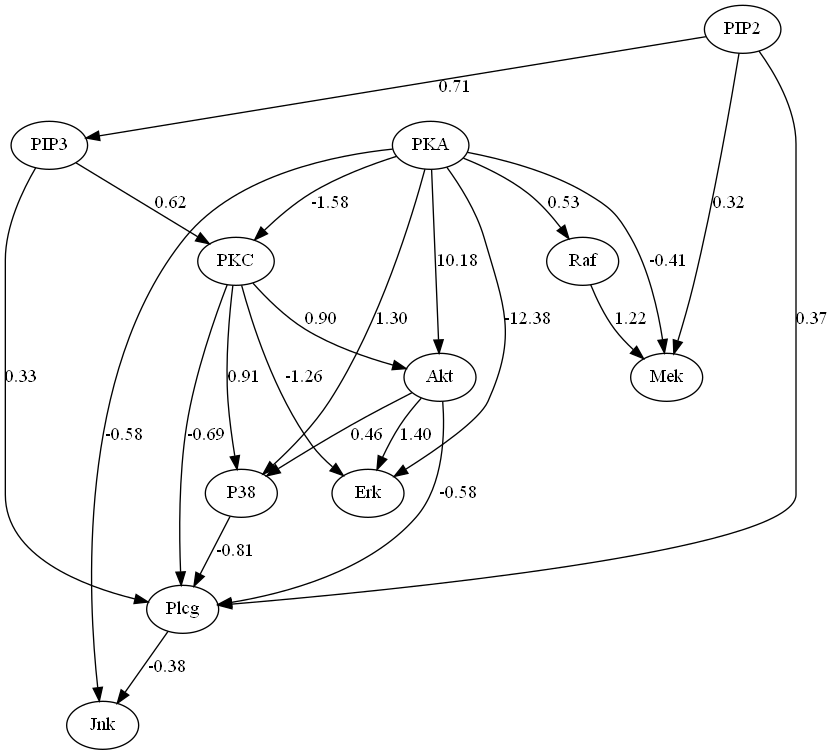}
        \\(c) $\epsilon=0.75$
    \end{minipage}
    \hfill
    \begin{minipage}{0.45\textwidth}
        \centering
        \includegraphics[width=\linewidth]{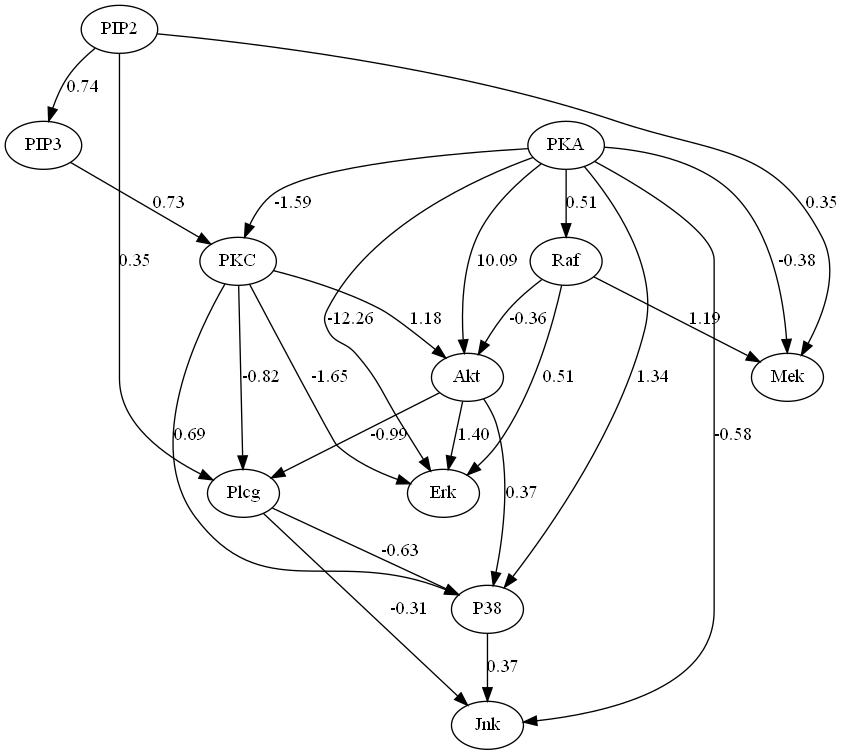}
        \\(d) $\epsilon=1.00$
    \end{minipage}
    
    \caption{Sachs causal models learned by Lin-CDIC (with interventional constraints) under different $\epsilon$ values.}
    \label{fig:sachs_models_cdic_all}
\end{figure}

\subsubsection{Effectiveness Analysis}
\noindent\textbf{Setting}: To demonstrate the effectiveness of interventional constraints, we use only the observational Sachs data ($n = 853$ samples) along with three of the eight identified interventional constraints: “PKC activates Jnk,” “PKC activates P38,” and “PIP3 activates Akt,” reserving the remaining five for validation. Accordingly, for \textbf{Lin-CD-Path} method that incorporates path constraints, the corresponding path constraints are: ``PKC $\to \dots \to$ Jnk'', ``PKC $\to \dots \to$ P38'', and ``PIP3 $\to \dots \to$ Akt''. The true causal graph and the causal models learned by \textbf{NOTEARS} (without constraints), \textbf{Lin-CD-Path} (with path constraints), and \textbf{Lin-CDIC} (with interventional constraints) for $\epsilon = 0.25$, $0.50$, $0.75$, and $1.0$ are shown in Figures~\ref{fig:sachs_model_notears_lin_cd_path}--\ref{fig:sachs_models_cdic_all}. The total causal effects of variable pairs and the performance metrics of the learned models are presented in Table \ref{tab:Effects_metrics_effectiveness}. Better metrics are shown in bold and blue. Note that in previous synthetic experiments, the signs of elements in the weight matrices are known, enabling evaluation of the learned models using the SCS metric. In contrast, for the real-world Sachs dataset, the signs and underlying cellular signalling mechanisms are only partially understood, making the SCS metric inapplicable for evaluation. Nevertheless, the signs in the learned model can still be verified against known causal interactions.

\noindent\textbf{Analysis:} From Table \ref{tab:Effects_metrics_effectiveness}, we observe that the model learned by \textbf{NOTEARS} without constraints satisfies only one of eight interventional constraints, specifically ``Raf \textit{activates} Mek'' with $T(\text{Raf}, \text{Mek})=1.21$ and two causal paths: PKA $\to \dots \to$ P38, and PKA $\to \dots \to$ Raf. However, it fails to identify key interactions: ``PKC \textit{activates} Jnk'', ``PKC \textit{activates} P38'', and ``PIP3 \textit{activates} Akt''. It also fails to identify corresponding causal paths: PKC $\to \dots \to$ Jnk, PKC $\to \dots \to$ P38, and PIP $\to \dots \to$ Akt. The model learned by \textbf{Lin-CD-Path} method incorporating path constraints shows improvement. Specifically, the causal interactions ``Raf \textit{activates} Mek'' and ``PIP3 \textit{activates} Akt'', as well as the causal paths PKC $\to \dots \to$ Jnk, PKC $\to \dots \to$ P38, PKA $\to \dots \to$ P38 and PKA $\to \dots \to$ Raf, are recovered. However, it fails to recover the causal interactions ``PKC \textit{activates} Jnk'',  ``PKC \textit{activates} P38'', ``PKA \textit{inhibits} P38'' and ``PKA \textit{inhibits} Raf'', and instead incorrectly infers ``PKC \textit{inhibits} Jnk'',  ``PKC \textit{inhibits} P38'', ``PKA \textit{activates} P38'' and ``PKA \textit{activates} Raf''. The model learned by our \textbf{Lin-CDIC} method incorporating interventional constraints, shows significantly better performance. Specifically, it satisfies all three specified interventional constraints: ``PKC \textit{activates} Jnk'', ``PKC \textit{activates} P38'', and ``PIP3 \textit{activates} Akt'', in addition to ``Raf \textit{activates} Mek''. Notably, it also uncovers a novel but unspecified causal interaction, ``PKA \textit{inhibits} P38'' with $T(\text{PKA}, \text{P38})=-0.4$, which means that it revealed two additional causal interactions: ``Raf \textit{activates} Mek'' and ``PKA \textit{inhibits} P38''. This suggests that leveraging partial interactions allow our method to successfully identify new and correct causal interactions. Additionally, our method also recovers causal pathways: PKA $\to \dots \to$ Raf, Raf $\to \dots \to$ Erk, and Mek $\to \dots \to$ Erk. However, the causal effects $T(\text{Raf}, \text{Erk})=-0.02$ and $T(\text{Mek}, \text{Erk})=-0.01$ indicate weak \textit{negative} causal effects, slightly violating the unspecified interactions, ``Raf \textit{activates} Erk'' and ``Mek \textit{activates} Erk''. Furthermore, $T(\text{PKA}, \text{Raf})=0.589$ contradicts the expected interaction, as PKA is expected to \textit{inhibits} Raf. In experiments with different $\epsilon$ values, when $\epsilon=0.50$, in addition to the three given interventional constraints, our method still successfully recovers two additional interactions: ``Raf \textit{activates} Mek'' and ``PKA \textit{inhibits} P38.'' Specifically, $T(\text{PKA}, \text{P38})$ is -4.32, indicating a stronger negative causal effect from PKA to P38 compared to -0.40 when $\epsilon=0.25$. However, when $\epsilon=0.75$ and 1.0, only ``Raf \textit{activates} Mek'' is consistently recovered. The value of $T(\text{PKA}, \text{P38})$ shifts to 3.93 and 7.42, respectively, suggesting ``PKA \textit{activates} P38,'' which contradicts the true interaction. Despite this inconsistency, our method still recovers the causal path from PKA to P38. The discrepancy among the four $\epsilon$ settings can likely be attributed to significant structural and parametric changes in the models caused by larger $\epsilon$ values. This observation aligns with our sensitivity analysis, where $\epsilon=0.25$ is found to be optimal among the four tested choices. In summary, given three interventional constraints/interactions, \textbf{Lin-CDIC} recovers two additional causal interactions (``Raf \textit{activates} Mek'' and ``PKA \textit{inhibits} P38''), and identifies five additional causal paths (PKC $\to \dots \to$ Jnk, PKC $\to \dots \to$ P38, PIP $\to \dots \to$ Akt, Raf $\to \dots \to$ Erk, and Mek $\to \dots \to$ Erk). These findings suggest that interventional constraints are more effective than path constraints, as correctly identifying causal interactions requires determining both the correct path and the appropriate sign of the causal effect. Additionally, interventional constraints on local causal interactions can, to some extent, facilitate the broader identification of causal interactions or paths. In addition, the causal models learned by \textbf{Lin-CDIC} with $\epsilon = 0.25$, $0.50$, and $0.75$ contain 22 edges, aligning more closely with the benchmark causal graph in Figure~\ref{fig:true_sachs_graph}, which has 20 edges, than those learned by \textbf{NOTEARS} and \textbf{Lin-CD-Path}. It is worth noting that in the real-world Sachs dataset experiment, although the sample size of 853 is relatively larger than those in the synthetic experiments, the performance metrics—such as FDR, TPR, FPR, SHD, and SID—of causal models estimated with or without constraints remain suboptimal. This may be attributed to measurement errors, noise, and unobserved confounders inherent in real-world data, which often require larger sample sizes for reliable causal discovery. In such scenarios, incorporating domain knowledge, such as interventional constraints, becomes essential.

\begin{table*}[ht]
\centering
\footnotesize
\setlength{\tabcolsep}{4pt}
\renewcommand{\arraystretch}{0.9}
\begin{threeparttable}
\begin{tabular}{|c|c|c|C{1.2cm}|C{1.2cm}|C{1.2cm}|C{1.2cm}|}
\hline
\textbf{\tiny Effect/Metrics} 
& \textbf{\tiny NOTEARS} 
& \textbf{\tiny Lin-CD-Path} 
& \textbf{\tiny Lin-CDIC} $\epsilon{=}0.25$ 
& \textbf{\tiny Lin-CDIC} $\epsilon{=}0.50$ 
& \textbf{\tiny Lin-CDIC} $\epsilon{=}0.75$ 
& \textbf{\tiny Lin-CDIC} $\epsilon{=}1.0$ \\
\hline
$T(\text{Raf}, \text{Mek}) > 0$ & \textbf{\textcolor{blue}{1.21}} & \textbf{\textcolor{blue}{1.21}} & \textbf{\textcolor{blue}{1.29}} & \textbf{\textcolor{blue}{1.21}} & \textbf{\textcolor{blue}{1.22}} & \textbf{\textcolor{blue}{1.19}} \\ \hline
$T(\text{PKC}, \text{Jnk}) > 0$ & 0 & -0.62 & \textbf{\textcolor{blue}{0.36}} & \textbf{\textcolor{blue}{0.43}} & \textbf{\textcolor{blue}{0.88}} & \textbf{\textcolor{blue}{1.51}} \\ \hline
$T(\text{PKC}, \text{P38}) > 0$ & 0 & -0.62 & \textbf{\textcolor{blue}{1.16}} & \textbf{\textcolor{blue}{1.41}} & \textbf{\textcolor{blue}{1.33}} & \textbf{\textcolor{blue}{2.39}} \\ \hline
$T(\text{PIP3}, \text{Akt}) > 0$ & 0 & \textbf{\textcolor{blue}{0.62}} & \textbf{\textcolor{blue}{0.41}} & \textbf{\textcolor{blue}{0.52}} & \textbf{\textcolor{blue}{0.56}} & \textbf{\textcolor{blue}{0.86}} \\ \hline
$T(\text{PKA}, \text{P38}) < 0$ & 0.96 & 2.28 & \textbf{\textcolor{blue}{-0.40}} & \textbf{\textcolor{blue}{-4.32}} & 3.93 & 7.42 \\ \hline
$T(\text{PKA}, \text{Raf}) < 0$ & 0.52 & 0.52 & 0.59 & 0.51 & 0.53 & 0.51 \\ \hline
$T(\text{Raf}, \text{Erk}) > 0$ & 0 & 0 & -0.02 & 0 & 0 & -0.00 \\ \hline
$T(\text{Mek}, \text{Erk}) > 0$ & 0 & 0 & -0.01 & 0 & 0 & 0 \\ \hline
FDR & 0.53 & \textbf{\textcolor{blue}{0.38}} & 0.50 & 0.64 & 0.64 & 0.67 \\ \hline
TPR & 0.35 & 0.50 & \textbf{\textcolor{blue}{0.55}} & 0.40 & 0.40 & 0.40 \\ \hline
FPR & 0.23 & \textbf{\textcolor{blue}{0.17}} & 0.31 & 0.40 & 0.40 & 0.46 \\ \hline
SHD & 14 & \textbf{\textcolor{blue}{11}} & 15 & 21 & 20 & 23 \\ \hline
SID & 47 & 38 & \textbf{\textcolor{blue}{31}} & \textbf{\textcolor{blue}{31}} & 35 & 34 \\ \hline
NNZ & 15 & 16 & \textbf{\textcolor{blue}{22}} & \textbf{\textcolor{blue}{22}}& \textbf{\textcolor{blue}{22}} & 24 \\ \hline
Time (s) & \textbf{\textcolor{blue}{2}} & 154 & 575 & 509 & 538 & 499 \\
\hline
\end{tabular}
\caption{Total causal effects and evaluation metrics of the causal models learned without constraints, with path constraints, and with interventional constraints under different $\epsilon$.}
\label{tab:Effects_metrics_effectiveness}
\end{threeparttable}
\end{table*}

\noindent\textbf{Remark:} Nonlinear models generally outperform linear models in causal discovery tasks. For example, on the Sachs dataset using purely observational data, nonlinear methods such as SCORE (\textcolor{blue}{\cite{rolland2022score}}) (SHD: 12, SID: 45), CAM (\textcolor{blue}{\cite{buhlmann2014cam}}) (SHD: 12, SID: 55), DiffAN (\textcolor{blue}{\cite{sanchez2023diffusion}}) (SHD: 13, SID: 56), and GraN-DAG (\textcolor{blue}{\cite{lachapelle2020gradient}}) (SHD: 13, SID: 47) have demonstrated superior performance, as reported by \textcolor{blue}{\cite{sanchez2023diffusion}}. In contrast, linear models like NOTEARS and FGS tend to yield higher Structural Hamming Distances (\textcolor{blue}{\cite{NOTEARS}} and \textcolor{blue}{\cite{yu2019daggnn}}). Although our method assumes a linear causal model, the SID metric value of the learned causal model, achieved using only three interventional constraints, is much lower than that of causal models learned under a nonlinear assumption.

\subsubsection{Robustness Analysis}
\noindent\textbf{Setting}: We also conducted a robustness analysis of our \textbf{Lin-CDIC} method. Specifically, we re-learned the causal models under the following combinations of interventional constraints: (1) one incorrect (“PIP3 \textit{inhibits} Akt”) and two correct (“PKC \textit{activates} Jnk”, “PKC \textit{activates} P38”); (2) two incorrect (“PIP3 \textit{inhibits} Akt”, “PKC \textit{inhibits} P38”) and one correct (“PKC \textit{activates} Jnk”); and (3) three incorrect constraints (“PIP3 \textit{inhibits} Akt”, “PKC \textit{inhibits} P38”, and “PKC \textit{inhibits} Jnk”). These results are compared with models learned by \textbf{NOTEARS} (without any constraints), \textbf{Lin-CD-Path} (with path constraints), and \textbf{Lin-CDIC} (with all correct interventional constraints). The total causal effects of variable pairs and the performance metrics of the learned models are presented in Table \ref{tab:Effects_metrics_robustness}. Note that \textbf{Lin-CD-Path} is not affected by the signs of causal effects or the correctness of interventional constraints. For example, for \textbf{Lin-CD-Path}, both “PIP3 \textit{inhibits} Akt” and “PIP3 \textit{activates} Akt” imply the existence of a causal path from PIP3 to Akt, i.e., PIP3 $\to \dots \to$ Akt. 

\noindent\textbf{Analysis:} Table \ref{tab:Effects_metrics_robustness} shows that introducing incorrect interventional constraints or priors results in sparser learned causal models. For example, when $\epsilon = 0.25$, the NNZ metric decreases from 22 to 20, indicating that two causal paths are missing compared to the model trained with all correct interventional constraints. Moreover, the incorrect constraints negatively influence the correct ones. For instance, when the incorrect constraint 'PIP3 \textit{inhibits} Akt' is provided, the causal path from PKC to Jnk becomes significantly weaker (e.g., 0.00 and -0.000), in contrast to the value of 0.36 obtained when all constraints are correct. This aligns with the earlier observation that incorporating incorrect constraints tends to produce sparser causal models. Among the models trained without constraints and with 0 to 3 correct interventional constraints, the combination of two incorrect and one correct constraint yields the best performance in terms of FDR, FPR, and SHD. This may be attributed to the relatively sparse model learned under that setting, as sparser models tend to exhibit fewer false edges. Interestingly, even when the signs of the interventional constraints are incorrect, they may still indicate correct causal paths, thereby improving structural metrics such as FDR, FPR, and SHD. This also highlights the effectiveness of our \textbf{Lin-CDIC} method in incorporating causal path priors, a topic that has been explored in prior work. In contrast, the model learned with all correct interventional constraints performs best on the SID metric, which evaluates the model from a downstream causal inference perspective. In addition, the causal models learned by \textbf{Lin-CDIC} contain between 16 and 22 edges, aligning more closely with the benchmark causal graph, which has 20 edges, than those learned by \textbf{NOTEARS} and \textbf{Lin-CD-Path}.

\begin{table*}[ht]
\centering
\footnotesize
\setlength{\tabcolsep}{4pt}
\renewcommand{\arraystretch}{0.9}
\begin{threeparttable}
\begin{tabular}
{|c|c|c|C{1.2cm}|C{1.2cm}|C{1.2cm}|C{1.2cm}|}
\hline
\textbf{\tiny Effect/Metrics} 
& \textbf{\tiny NOTEARS} 
& \textbf{\tiny Lin-CD-Path} 
& \textbf{\tiny Lin-CDIC} \textbf{\tiny IC-3} 
& \textbf{\tiny Lin-CDIC} \textbf{\tiny IC-2} 
& \textbf{\tiny Lin-CDIC} \textbf{\tiny IC-1} 
& \textbf{\tiny Lin-CDIC} \textbf{\tiny IC-0} \\
\hline
$T(\text{Raf}, \text{Mek}) > 0$      & \textbf{\textcolor{blue}{1.21}} & \textbf{\textcolor{blue}{1.21}} & \textbf{\textcolor{blue}{1.21}} & \textbf{\textcolor{blue}{1.22}} & \textbf{\textcolor{blue}{1.20}} & \textbf{\textcolor{blue}{1.29}} \\ \hline
$T(\text{PKC}, \text{Jnk}) > 0$      & 0 & -0.62 & -0.00 & \textbf{\textcolor{blue}{0.37}} & \textbf{\textcolor{blue}{0.00}} & \textbf{\textcolor{blue}{0.36}} \\ \hline
$T(\text{PKC}, \text{P38}) > 0$      & 0 & -0.62 & -0.45 & -0.49 & \textbf{\textcolor{blue}{0.57}} & \textbf{\textcolor{blue}{1.16}} \\ \hline
$T(\text{PIP3}, \text{Akt}) > 0$     & 0 & \textbf{\textcolor{blue}{0.62}} & -0.66 & -0.43 & -0.43 & \textbf{\textcolor{blue}{0.41}} \\ \hline
$T(\text{PKA}, \text{P38}) < 0$      & 0.96 & 2.28 & 2.06 & 2.05 & 0.31 & \textbf{\textcolor{blue}{-0.40}} \\ \hline
$T(\text{PKA}, \text{Raf}) < 0$      & 0.52 & 0.52 & 0.51 & 0.52 & 0.52 & 0.59 \\ \hline
$T(\text{Raf}, \text{Erk}) > 0$      & 0 & 0 & 0 & 0 & 0 & -0.02 \\ \hline
$T(\text{Mek}, \text{Erk}) > 0$      & 0 & 0 & 0 & 0 & 0 & -0.01 \\ \hline
FDR   & 0.53 & \textbf{\textcolor{blue}{0.38}} & 0.60 & \textbf{\textcolor{blue}{0.38}} & 0.44 & 0.50 \\ \hline
TPR   & 0.35 & 0.50 & 0.40 & 0.50 & 0.45 & \textbf{\textcolor{blue}{0.55}} \\ \hline
FPR   & 0.23 & \textbf{\textcolor{blue}{0.17}} & 0.34 & \textbf{\textcolor{blue}{0.17}} & 0.20 & 0.31 \\ \hline
SHD   & 14   & \textbf{\textcolor{blue}{11}} & 18 & \textbf{\textcolor{blue}{11}} & 12 & 15 \\ \hline
SID   & 47   & 38 & 35 & 38 & 43 & \textbf{\textcolor{blue}{31}} \\ \hline
NNZ   & 15   & 16 & \textbf{\textcolor{blue}{20}} & 16 & 16 & 22 \\ \hline
Time (s) & \textbf{\textcolor{blue}{2}} & 154 & 724 & 1493 & 675 & 575 \\
\hline
\end{tabular}
\vspace{0.5ex}
\begin{minipage}{0.95\textwidth}
\footnotesize
\textbf{Note:} \texttt{IC-n} denotes interventional constraints containing \textit{n} incorrect specifications. Bold values indicate total causal effects aligned with the ground truth or the best performance across metrics.

\end{minipage}
\vspace{1ex}
\caption{Total causal effects and evaluation metrics of the causal models learned without constraints, with path constraints, and with 0 to 3 correct interventional constraints.}
\label{tab:Effects_metrics_robustness}
\end{threeparttable}
\end{table*}

\subsubsection{Generalization Analysis}

\noindent\textbf{Setting}: We further analyzed the generalization of our method by cross-validating the interventional constraints. Based on Table \ref{tab:interventional_constraints}, there are $\binom{8}{3} = 56$ possible combinations of training constraint sets. We performed causal discovery for each combination using the corresponding path and interventional constraints. The average total causal effects of variable pairs and evaluation metrics of the causal models learned without constraints, with path constraints, and with interventional constraints are presented in Table \ref{tab:Effects_metrics_generalization}.

\begin{table*}[ht]
\centering
\footnotesize
\setlength{\tabcolsep}{4pt}
\renewcommand{\arraystretch}{0.9}
\begin{threeparttable}
\begin{tabular}{|c|c|c|C{1.2cm}|C{1.2cm}|C{1.2cm}|C{1.2cm}|}
\hline
\textbf{\tiny Effect/Metrics} 
& \textbf{\tiny NOTEARS} 
& \textbf{\tiny Lin-CD-Path} 
& \textbf{\tiny Lin-CDIC} $\epsilon{=}0.25$ 
& \textbf{\tiny Lin-CDIC} $\epsilon{=}0.50$ 
& \textbf{\tiny Lin-CDIC} $\epsilon{=}0.75$ 
& \textbf{\tiny Lin-CDIC} $\epsilon{=}1.0$ \\
\hline
$T(\text{Raf}, \text{Mek}) > 0$      & \textbf{\textcolor{blue}{1.21}} & \textbf{\textcolor{blue}{1.17}} & \textbf{\textcolor{blue}{1.25}} & \textbf{\textcolor{blue}{1.26}} & \textbf{\textcolor{blue}{1.42}} & \textbf{\textcolor{blue}{1.24}} \\ \hline
$T(\text{PKC}, \text{Jnk}) > 0$      & 0      & -0.13 & \textbf{\textcolor{blue}{0.18}} & \textbf{\textcolor{blue}{0.27}} & \textbf{\textcolor{blue}{0.41}} & \textbf{\textcolor{blue}{0.51}} \\ \hline
$T(\text{PKC}, \text{P38}) > 0$      & 0      & \textbf{\textcolor{blue}{0.19}} & \textbf{\textcolor{blue}{0.35}} & \textbf{\textcolor{blue}{0.46}} & \textbf{\textcolor{blue}{0.40}} & \textbf{\textcolor{blue}{0.65}} \\ \hline
$T(\text{PIP3}, \text{Akt}) > 0$     & 0      & -0.12 & \textbf{\textcolor{blue}{0.14}} & \textbf{\textcolor{blue}{0.20}} & \textbf{\textcolor{blue}{0.20}} & \textbf{\textcolor{blue}{0.37}} \\ \hline
$T(\text{PKA}, \text{P38}) < 0$      & 0.96   & 0.86  & \textbf{\textcolor{blue}{-0.94}} & \textbf{\textcolor{blue}{-1.95}} & \textbf{\textcolor{blue}{-2.59}} & \textbf{\textcolor{blue}{-7.08}} \\ \hline
$T(\text{PKA}, \text{Raf}) < 0$      & 0.52   & 0.47  & \textbf{\textcolor{blue}{-0.54}} & \textbf{\textcolor{blue}{-0.61}} & \textbf{\textcolor{blue}{-1.02}} & \textbf{\textcolor{blue}{-1.51}} \\ \hline
$T(\text{Raf}, \text{Erk}) > 0$      & 0      & \textbf{\textcolor{blue}{0.18}} & \textbf{\textcolor{blue}{0.40}} & \textbf{\textcolor{blue}{0.20}} & \textbf{\textcolor{blue}{0.43}}  & \textbf{\textcolor{blue}{0.66}} \\ \hline
$T(\text{Mek}, \text{Erk}) > 0$      & 0      & \textbf{\textcolor{blue}{0.13}}  & \textbf{\textcolor{blue}{0.29}} & \textbf{\textcolor{blue}{0.24}} & \textbf{\textcolor{blue}{0.53}}    & \textbf{\textcolor{blue}{0.49}}    \\ \hline
FDR   & 0.53 & \textbf{\textcolor{blue}{0.45}} & 0.59 & 0.61 & 0.64 & 0.62 \\ \hline
TPR   & 0.35 & \textbf{\textcolor{blue}{0.43}} & 0.41 & 0.42 & 0.40 & \textbf{\textcolor{blue}{0.43}} \\ \hline
FPR   & 0.23 & \textbf{\textcolor{blue}{0.20}} & 0.35 & 0.39 & 0.40 & 0.41 \\ \hline
SHD   & 14 & \textbf{\textcolor{blue}{13.14}} & 17.75 & 19.12 & 20.5 & 20.1 \\ \hline
SID   & 47 & 41.59 & 42.71 & 42.80 & 42.0 & \textbf{\textcolor{blue}{41.3}} \\ \hline
NNZ   & 15 & 15.59 & \textbf{\textcolor{blue}{20.3}} & 22.16 & 22.1 & 22.9 \\ \hline
Time (s)  & \textbf{\textcolor{blue}{2}} & 152 & 1028 & 833 & 706 & 501 \\
\hline
\end{tabular}
\caption{Average total causal effects and evaluation metrics of the learned causal models without constraints, with path constraints, and with interventional constraints ($\epsilon=0.25,0.50,0.75,1.0$).}
\label{tab:Effects_metrics_generalization}
\end{threeparttable}
\end{table*}
\noindent\textbf{Analysis:} Table \ref{tab:Effects_metrics_generalization} shows that, under three random constraints, the average total causal effects between variable pairs learned by our \textbf{Lin-CDIC} method remain consistent with previously established findings. In contrast, the results from \textbf{Lin-CD-Path} and \textbf{NOTEARS} align only partially, capturing a limited subset of known causal interactions. 1) In terms of the average metrics FDR, FPR, and SHD, the models learned by \textbf{Lin-CDIC} exhibit higher values compared to those learned by \textbf{Lin-CD-Path} and \textbf{NOTEARS}. This may be due to the higher density of the causal models produced by \textbf{Lin-CDIC}, which contain between 20.3 and 22.9 edges—denser than those from \textbf{NOTEARS} and \textbf{Lin-CD-Path}. Greater density can lead to more false positives, thereby increasing FDR, FPR, and SHD. 2) In terms of the average SID metric, the models learned by the \textbf{Lin-CDIC} method show slightly lower SID values at $\epsilon = 1.0$, and slightly higher values at $\epsilon = 0.25, 0.50,$ and $0.75$, compared to those learned by the \textbf{Lin-CD-Path} method. This variation may arise from uncertainties in the correctness of the assumed ground truth structure shown in Figure \ref{fig:true_sachs_graph}. For instance, \textcolor{blue}{\citealp{PIP3-Akt-2021}} suggest that Akt may indirectly inhibit further PIP3 synthesis through a feedback mechanism, implying a potential feedback loop between PIP3 and Akt, PIP3 $\to \dots \to$ Akt $\to \dots \to$ PIP3, an interaction not captured in the ground truth. \textcolor{blue}{\citealp[p. 10]{Sachs2009}} noted that the T-cell signaling pathway was believed to contain at least two feedback cycles—specifically, a longer loop Raf $\to$ Mek $\to$ Erk $\to$ Akt $\to$ Raf and a shorter loop Raf $\to$ Mek $\to$ Erk $\to$ Raf. \textcolor{blue}{\citealp{brouillard2024landscape}} revisited the Sachs dataset in a comprehensive review of causal discovery benchmarks and updated the “ground truth” graph to include a prominent feedback loop Raf $\to$ Mek $\to$ Erk $\to$ Raf (see their Figure~7). However, due to the acyclicity assumption adopted in this paper, we do not use their graph as the benchmark. It is worth noting that \textcolor{blue}{\citealp[p.~34]{brouillard2024landscape}} also advocate evaluating not only structure recovery but also interventional predictions, which reinforces the motivation of our study. Regarding the SID metric, it quantifies the number of inconsistencies between two causal graphs by comparing their resulting post-intervention distributions $P(Y \mid \text{do}(X))$ under all possible single-variable interventions. Intuitively, it captures the number of mismatches in causal pathways between the graphs. For instance, if the causal model learned by \textbf{Lin-CDIC} includes a causal path that is absent in the benchmark graph (Figure~\ref{fig:true_sachs_graph}), it is considered one inconsistency in the SID computation, thereby increasing the SID value for \textbf{Lin-CDIC}. Consequently, if the assumed ground-truth structure is uncertain, the SID value becomes equally unreliable. By relaxing the acyclicity assumption, \textbf{Lin-CDIC} may therefore achieve better performance on the Sachs dataset (see Discussion). 3) In terms of time consumption, \textbf{Lin-CDIC} exhibits a clear decrease as $\epsilon$ increases. This trend can be attributed to the nature of updates during optimization: smaller $\epsilon$ values lead to more conservative changes in the causal model, requiring more iterations to satisfy the given constraints. In contrast, larger $\epsilon$ values (e.g., $\epsilon = 0.75$ and $\epsilon = 1.0$) introduce more substantial updates, enabling the model to satisfy constraints more quickly. However, these larger updates may also risk underfitting or missing the optimal solution due to overly aggressive changes. Note that, due to the complexity of the optimization process, we did not conduct experiments using all interventional constraints. The primary reason is that when the number of interventional constraints exceeds five, \textbf{Lin-CDIC} often converges to a local optimum. We leave this limitation as an open direction for future research.

\section{Discussion}
We introduce interventional constraints, a novel causal knowledge concept, to enhance the accuracy and explainability of causal discovery. Empirical results show that these constraints not only enforce consistency with known findings but also uncover additional correct interactions and pathways. Future directions include: (1) Scalability remains a key challenge due to the high non-convexity and constraint burden. Future work will explore more efficient optimization strategies to support larger causal systems. (2) Extending linear causal discovery with interventional constraints in the presence of hidden confounders by integrating them with differentiable algebraic equality constraints that fully characterize ancestral ADMGs, as well as more general classes such as arid ADMGs and bow-free ADMGs (\textcolor{blue}{\cite{Linear_hidden}}). Since all these constraints are differentiable, they can be unified into a single framework. (3) Generalization to nonlinear models, where causal effect value depends on intervention values (\textcolor{blue}{\cite{Pearl-causal-effect-nonlinear}}) and may require neural network parameterizations (\textcolor{blue}{\cite{Kevin-Xia-nueral-causal}}). In these settings, optimizing path-specific effects calculated through nested functions can be challenging when multiple causal paths exist. (4) Incorporating interventional constraints into cyclic Structural Causal Models (SCMs) (\textcolor{blue}{\cite{hyttinen12a,Mooij-Sachs-2,Mooij-Sachs-1,Cycle-SCM, dai2024local}}) to create a more comprehensive framework for causal discovery in dynamic systems, such as biological systems, improving the ability to handle feedback loops and cyclic dependencies in real-world settings. (5) Decomposing Total Effects into Direct and Indirect Components. To assess global satisfaction of interventional constraints, we use the total causal effect, which captures both direct and indirect influences. While this provides a holistic measure, it may mask the contributions of specific causal pathways. Future work could enhance interpretability by explicitly separating direct and indirect effects. (6) Leveraging large language models (LLMs) to automatically extract high-level causal knowledge, enhancing scalability and explainability. While expert validation remains important (\textcolor{blue}{\cite{LLM_drawbacks}}), recent work demonstrates the potential of LLMs in guiding causal discovery (\textcolor{blue}{\cite{LLM_for_Causal_discovery_2023_1, LLM-for-causal-discovery-2024, LLM-Causality-Survey-2024-1, LLM_ordering-2023, LLM-query-2023}}), making them a promising addition to our framework.

\section*{Acknowledgments}

This work was supported by the UK Engineering and Physical Sciences Research Council (EPSRC) under grant number EP/X029778/1, titled \textit{“Causal Counterfactual Visualisation for Human Causal Decision Making – A Case Study in Healthcare”}.
\bibliography{references}

\newpage
\begin{appendices}
\begin{center}
  \Large \bfseries Appendix: Linear Causal Discovery with Interventional Constraints
\end{center}

\section{SLSQP Algorithm} \label{sec:slsqp_algorithm}
\begin{algorithm}[htbp]
\caption{SLSQP Algorithm}
\label{alg:slsqp}
\begin{algorithmic}[1]
\Require Initial weight matrix $W^{(1)}$, data $X$, constraint thresholds $\delta$, interventional constraints $\mathcal{I}$, objective function $F(W)$, gradient $\nabla F(W)$, causal effect measure $T(W)$, acyclicity measure $h(W)$, bounds on variables $\mathcal{B}$, maximum iterations $max\_iter$, tolerance $tol$
\Ensure Estimated weight matrix $W_{\text{est}}$
\State $k \gets 0$
\State $W \gets W^{(1)}$
\State $h_{\text{val}} \gets h(W)$
\State $F_{\text{val}} \gets F(W)$
\State $T_{\text{val}} \gets T(W, \mathcal{I})$
\State $J_T \gets$ Jacobian of $T(W)$
\State $J_h \gets$ Jacobian of $h(W)$
\State $convergence\_flag \gets \text{False}$
\While{True}
    \Comment{Solve quadratic subproblem:}
    \State Minimize $\nabla F(W)^T \Delta W + \frac{1}{2} \Delta W^T H \Delta W$
    \State Subject to:
    \State \hspace{1em} $J_T \Delta W + T_{\text{val}} \leq 0$
    \State \hspace{1em} $J_h \Delta W + h_{\text{val}} = 0$
    \State \hspace{1em} $\Delta W \in \mathcal{B}$
    \State $W_{\text{est}} \gets W + \Delta W$
    \If{$\|W_{\text{est}} - W\| < tol$}
        \State $convergence\_flag \gets \text{True}$
    \EndIf
    \State $k \gets k + 1$
    \State $W \gets W_{\text{est}}$
    \State $h_{\text{val}} \gets h(W)$
    \State $F_{\text{val}} \gets F(W)$
    \State $T_{\text{val}} \gets T(W, \mathcal{I})$
    \State $J_T \gets$ Jacobian of $T(W)$
    \State $J_h \gets$ Jacobian of $h(W)$
    \If{$convergence\_flag$ \textbf{or} $k \geq max\_iter$}
        \State \textbf{break}
    \EndIf
\EndWhile
\State \Return $W_{\text{est}}$
\end{algorithmic}
\end{algorithm}

\section{Linear Causal Discovery with Path Constraint Algorithm} \label{sec:cd_path_algorithm} 
In this paper, we implement the causal discovery algorithm with general path constraints, similar to our proposed Lin-CDIC algorithm. The main difference lies in replacing the interventional constraints $\delta_{ij}\bigl(T_{i,j} - \delta_{ij}\bigr) > 0,\ i \in \mathcal{C},\ j \in \mathcal{T}$ with reachability (or path-based) constraints $\bigl(R_{i,j} - \rho_{ij}\bigr) > 0,\ i \in \mathcal{C},\ j \in \mathcal{T}$, where $R_{ij}$ is defined as below:

\[
R = \left(I + \frac{\tanh(W)}{d}\right)^d,
\]
where $W \in \mathbb{R}^{d \times d}$, $d>0$ is the number of variables.

\begin{tcolorbox}[colback=white, colframe=black, title=\textbf{Proposition B.1}]
The absolute value of any entry of \( R \) satisfies  
\[
\max_{i,j} |R_{ij}| = \frac{2^d - 1}{d}.
\]
\end{tcolorbox}

\begin{proof}: 
For any real number $W_{ij}$, $\tanh(W_{ij}) \in (-1, 1)$. Thus, the elements of $\frac{\tanh(W)}{d}$ lie in $\left(-\frac{1}{d}, \frac{1}{d}\right)$. Consider the case where all elements of $W \to +\infty$, so $\tanh(W) \to 1$. Then,
\[
M = I + \frac{1}{d} \mathbf{1},
\]
where $\mathbf{1}$ is the all-ones matrix. Using the binomial expansion,
\[
M^d = \sum_{k=0}^{d} \binom{d}{k} I^{d-k} \left(\frac{1}{d} \mathbf{1} \right)^k.
\]
Noting that $\mathbf{1}^k = d^{k-1} \mathbf{1}$ for $k \geq 1$, we have
\[
M^d = I + \frac{1}{d} \left(2^d - 1\right) \mathbf{1}.
\]
Therefore, 

\[
R_{ij} = 
\begin{cases} 
1 & \text{if } i = j \quad (\text{diagonal entries}) \\[1ex]
\displaystyle \frac{2^d - 1}{d} & \text{if } i \neq j \quad (\text{maximum off-diagonal value}) 
\end{cases}
\]
Hence, 
\[
\max_{i,j} |R_{ij}| = \frac{2^d - 1}{d}.
\]
Consider the case where all elements of $W \to -\infty$, so $\tanh(W) \to -1$. Then, 
\[
M = I - \frac{1}{d} \mathbf{1}.
\]
Applying the binomial expansion again:
\[
M^d = I + \frac{1}{d} \left((-1)^d (2^d - 1)\right) \mathbf{1}.
\]
If $d$ is even, the off-diagonal entries remain positive:
\[
R_{ij} = \frac{2^d - 1}{d}.
\]
If $d$ is odd, the off-diagonal entries become negative:
\[
R_{ij} = -\frac{2^d - 1}{d}.
\]
In both cases, considering the absolute value, 
\[
\max_{i,j} |R_{ij}| = \frac{2^d - 1}{d}.
\]
Therefore, the maximum absolute value of any entry of $R$ is exactly $\frac{2^d - 1}{d}$, regardless of the values of $W$.

\[
\boxed{\max_{i,j} |R_{ij}| = \frac{2^d - 1}{d} \quad \text{(for } i \neq j \text{)}}
\]
\end{proof}

\begin{tcolorbox}[colback=white, colframe=black, title=\textbf{Proposition B.2}]
$R$ is most sensitive to $|W_{ij}|$) when $W_{ij}$ is near zero, whereas as $|W_{ij}|$ becomes large, its effect on $R$ becomes negligible.
\end{tcolorbox}

\begin{proof}: 

\[
\frac{\partial R}{\partial |W_{ij}|} = \frac{\partial R}{\partial M} \cdot \frac{\partial M}{\partial \tanh(W_{ij})} \cdot \frac{\partial \tanh(W_{ij})}{\partial W_{ij}} \cdot \frac{\partial W_{ij}}{\partial |W_{ij}|}.
\]

\noindent Since $R = M^d$, 
    \[
    \frac{\partial R}{\partial M} = d M^{d-1}.
    \]

\noindent From $M = I + \frac{\tanh(W)}{d}$, we have 
    \[
    \frac{\partial M}{\partial \tanh(W_{ij})} = \frac{1}{d}.
    \]
    
\noindent The derivative of $\tanh(W_{ij})$ is 
    \[
    \frac{\partial \tanh(W_{ij})}{\partial W_{ij}} = 1 - \tanh^2(W_{ij}).
    \]
    
\noindent The derivative of $W_{ij}$ with respect to $|W_{ij}|$ is 
    \[
    \frac{\partial W_{ij}}{\partial |W_{ij}|} = \operatorname{sign}(W_{ij}).
    \]
    
\noindent Therefore,
\[
\frac{\partial R}{\partial |W_{ij}|} = d M^{d-1} \cdot \frac{1}{d} \cdot \left(1 - \tanh^2(W_{ij})\right) \cdot \operatorname{sign}(W_{ij}) = M^{d-1} \cdot \left(1 - \tanh^2(W_{ij})\right) \cdot \operatorname{sign}(W_{ij}).
\]

\noindent Extracting the $(i, j)$-th element, we have:
\[
\frac{\partial R}{\partial |W_{ij}|} = \left(M^{d-1}\right)_{ij} \cdot \left(1 - \tanh^2(W_{ij})\right) \cdot \operatorname{sign}(W_{ij}).
\]

\noindent From $\frac{\partial R}{\partial |W_{ij}|}$, we can conclude, when $|W_{ij}| \to 0$, we have $\tanh(W_{ij}) \to 0$, and hence $1 - \tanh^2(W_{ij}) \to 1$. Therefore, $R$ is most sensitive to $|W_{ij}|$.  When $|W_{ij}| \to \infty$, we have $\tanh(W_{ij}) \to \pm 1$, and $1 - \tanh^2(W_{ij}) \to 0$. Thus, sensitivity approaches zero.
\noindent\textbf{In summary, the sensitivity of 
$R$ to $|W_{ij}|$ is highest near zero and gradually diminishes as $|W_{ij}|$ increases.}
\end{proof}

In this paper, the default value of the weight threshold $\omega$ is set to 0.3. Therefore, we further analyze the sensitivity of $|W_{ij}|$ with respect to $|W_{ij}|$ when $|W_{ij}|$ varies around 0.3.

\begin{tcolorbox}[colback=white, colframe=black, title=\textbf{Proposition B.3}]
When $|W_{ij}|$ varies around 0.3, the sensitivity retains approximately \textbf{91.5\%} of its maximum value.
\end{tcolorbox}

\begin{proof}: 
The sensitivity of $R$ with respect to the absolute value $|W_{ij}|$ is given by
\[
\frac{\partial R}{\partial |W_{ij}|} = \left( M^{d-1} \right)_{ij} \cdot \left(1 - \tanh^2(W_{ij})\right) \cdot \operatorname{sign}(W_{ij}),
\]
where $M = I + \frac{\tanh(W)}{d}$. The critical sensitivity factor is $1 - \tanh^2(W_{ij}),$ which determines the sensitivity behavior as $W_{ij}$ changes. When $|W_{ij}| \approx 0.3$, $\tanh(0.3) \approx 0.291.$ Therefore,
    \[
    1 - \tanh^2(0.3) \approx 1 - (0.291)^2 \approx 0.915.
    \]
    
\noindent Therefore, when $|W_{ij}|$ varies around 0.3, the sensitivity retains approximately \textbf{91.5\%} of its maximum value. This indicates that, compared with changes in $|W_{ij}|$ when $|W_{ij}| \approx 0$, $R$ becomes less sensitive to changes in $|W_{ij}|$ around 0.3.
\end{proof}

In this paper, the path constraint $\bigl(R_{i,j} - \rho_{ij}\bigr) > 0,\ i \in \mathcal{C},\ j \in \mathcal{T}$ is also implemented in the two-stage optimization method. Since $R$ is highly sensitive to $|W_{ij}|$, we naively set $\epsilon$ to 0.01. The causal discovery with path constraints algorithm is summarized in Algorithm \ref{alg:cd_path}.

\begin{algorithm}[ht]
\caption{\textsc{Lin-CD-Path} Algorithm}
\label{alg:cd_path}
\begin{algorithmic}[1]
\Require Observational data $X$, cause variable set $\mathcal{C}$, target variable set $\mathcal{T}$, acyclicity tolerance $h_{tol}$, weight threshold $\omega$, adjustment factor $\epsilon$
\Ensure Optimal weight matrix $W^*$
\State $ConSat \gets \text{False}$ \Comment{Satisfaction of constraints}
\State $W^{(1)} \gets \text{L-BFGS-B}(X,\, h_{tol})$
\State $\rho \gets \{\rho_{ij} = 0 \mid i \in \mathcal{C}, j \in \mathcal{T}\}$ \Comment{Constraint thresholds}
\State $\mathcal{I} \gets \emptyset$ \Comment{Accumulated path constraints}
\For{each $i \in \mathcal{C}$ and $j \in \mathcal{T}$}
    \State $\mathcal{I} \gets \mathcal{I} \cup \{R_{ij} > 0\}$ \Comment{Add path constraints}
    \While{True}
        \State $W_{\text{est}} \gets \text{SLSQP}(F(W),\, X,\, W^{(1)},\, \rho,\, \mathcal{I})$
        \State $W^* \gets W_{\text{est}} \circ \mathbf{1}(|W_{\text{est}}| > \omega)$
        \State $ConSat \gets \text{Constraint\_check}(W^*,\, \mathcal{I})$
        \If{$W^*$ is a DAG}
            \If{$ConSat$ is \text{True}}
                \State $W_0 \gets W_{\text{est}}$
                \State \textbf{break}
            \Else
                \State $\rho_{ij} \gets \rho_{ij} + \epsilon$ \Comment{Threshold adjustment}
            \EndIf
        \Else
            \State $h_{tol} \gets h_{tol} \times 0.25$
        \EndIf
    \EndWhile
\EndFor
\State \Return $W^*$
\end{algorithmic}
\end{algorithm}

\section{Sensitivity Analysis of $\epsilon$ Values} \label{sec:epsilon}

\begin{tcolorbox}[colback=white, colframe=black, title=\textbf{Proposition C.1}]
$T = (I - W)^{-1} - I$ is significantly more sensitive to changes in $W_{ij}$ than 
\[
R = \left(I + \frac{\tanh(W)}{d}\right)^d.
\]
\end{tcolorbox}

\begin{proof}
The sensitivity of $T$ with respect to changes in $W_{pq}$ is given by  
\[
\frac{\partial T_{ij}}{\partial W_{pq}} = \left[(I - W)^{-1}\right]_{ip} \cdot \left[(I - W)^{-1}\right]_{qj}.
\]  
This expression shows that the sensitivity of $T$ depends on the entries of $(I - W)^{-1}$, which capture the cumulative effects of feedback loops in the system. As the entries of $W$ increase, $(I - W)^{-1}$ can grow rapidly, especially as $\| W \| \to 1$. This leads to potentially unbounded and exponentially increasing sensitivity, making $T$ highly unstable under changes in $W_{ij}$. The sensitivity of $R$ with respect to the absolute value $|W_{ij}|$ is  
\[
\frac{\partial R}{\partial |W_{ij}|} = \left( M^{d-1} \right)_{ij} \cdot \left(1 - \tanh^2(W_{ij})\right) \cdot \operatorname{sign}(W_{ij}),
\]  
where $M = I + \frac{\tanh(W)}{d}$. Among the terms on the right-hand side, note that $1 - \tanh^2(W_{ij}) = \text{sech}^2(W_{ij})$, which decreases rapidly as $|W_{ij}|$ increases. As $|W_{ij}| \to \infty$, $\tanh(W_{ij}) \to 1$, so $\text{sech}^2(W_{ij}) \to 0$, and the sensitivity of $R$ approaches zero. This saturation effect of the $\tanh$ function naturally limits the sensitivity of $R$, ensuring that changes in $W_{ij}$ have a bounded and diminishing influence on $R$.

\medskip

In general, the sensitivity of $T$ increases rapidly and can become unbounded as $W_{ij}$ grows, especially when the spectral norm $\| W \|$ approaches 1. In contrast, the sensitivity of $R$ remains bounded and decreases as $|W_{ij}|$ increases, due to the saturation behavior of the $\tanh$ function. Therefore, $T$ is significantly more sensitive to changes in $W_{ij}$ than $R$, and is much more prone to instability in response to perturbations of the matrix $W$.
\end{proof}

We empirically investigate how to choose $\epsilon$. Specifically, we generate random linear causal models characterized by scale-free (SF) graphs (\textcolor{blue}{\citealp{Scale-free}}) with Gaussian noise. The number of causal edges is also randomly determined, falling between eight and $\min\left(\left\lfloor \frac{d \cdot (d - 1)}{2} \right\rfloor, 10\right)$, where $d$ is the number of nodes. As for the interventional constraints, we sample from the true causal model based on the strength of the causal effects between cause and target variables. A causal effect from variable $i$ to $j$, denoted as $T_{ij}$, is considered significant if $|T_{ij}| > 0.1$ and is likely to be sampled. The above definition has real-world implications in fields such as genomics, econometrics, and systems biology. For example, weak causal effects are often seen as potentially spurious connections. We considered four $\epsilon$ value settings: $\epsilon = 0.25$, 0.5, 0.75, and 1.0, and demonstrated the effect of these $\epsilon$ values by testing 20 random experiments with 10 variables and a sample size of 100. In each experiment, we generated two interventional constraints. We selected the experiments where interventional constraints were violated. The performance of Lin-CDIC under different $\epsilon$ values is summarized in Table \ref{tab:epsilon}. Better metrics are shown in bold and blue.

\begin{table}[ht]
\centering
\footnotesize 
\setlength{\tabcolsep}{4pt} 
\renewcommand{\arraystretch}{0.85}
\begin{tabular}{@{}lcccc@{}}
\toprule
Metrics & $\epsilon=0.25$ & $\epsilon=0.5$ & $\epsilon=0.75$ & $\epsilon=1.0$ \\ 
\midrule
FDR    & (\textbf{\textcolor{blue}{0.116, 0.009}}) & (0.153, 0.020) & (0.123, 0.013) & (0.171, 0.023) \\
TPR    & (\textbf{\textcolor{blue}{0.856, 0.008}}) & (0.841, 0.008) & (0.852, 0.010) & (0.832, 0.011) \\
FPR    & (\textbf{\textcolor{blue}{0.058, 0.003}}) & (0.088, 0.015) & (0.059, 0.003) & (0.094, 0.015) \\
SHD    & (\textbf{\textcolor{blue}{3.200, 5.660}}) & (4.350, 19.828)  & (3.300, 8.310)  & (4.600, 20.040)  \\
SID    & (\textbf{\textcolor{blue}{4.450, 14.748}})& (5.650, 18.428)  & (4.550, 15.948)  & (5.850, 22.028)  \\
NNZ    & (14.300, 14.510)        & (14.900, 20.190) & (\textbf{\textcolor{blue}{14.250, 11.288}}) & (14.950, 18.248) \\
SCS    & \textbf{\textcolor{blue}{1,926}}   & 1,900 & 1,923 & 1,923 \\
Time   & 27.89 s  & 30.96 s & \textbf{\textcolor{blue}{27.87 s}}        & 27.89 s       \\
\bottomrule
\end{tabular}
\caption{Performance of Lin-CDIC under different $\epsilon$ values. The mean and variance of the edge numbers, i.e. NNZ, in the generated causal models are 13.05 and 12.25, respectively.}
\label{tab:epsilon}
\end{table}

\noindent\textbf{Accuracy:} From Table \ref{tab:epsilon}, we observe that metrics, including FDR, TPR, FPR, SHD, and SID, of estimated causal models under the setting $\epsilon = 0.25$ outperform those under other $\epsilon$ settings. This can be explained by the fact that smaller updates to $\epsilon$ result in slight changes to the causal model during the optimization process, while larger updates, such as $\epsilon = 0.75$ and $\epsilon = 1.0$, lead to more significant changes. When these changes are large, the learned models are likely to underfit. Conversely, we expect only minor changes—primarily adjustments to the existence and strength of causal paths between the cause and target variables constrained by the given interventions. This is evident in the causal models learned with 
$\epsilon = 0.25$, which are relatively sparser, as indicated by the number of non-zero elements (NNZ), i.e., the number of edges. Settings with $\epsilon = 1.0$ result in denser networks, which is also why we did not consider $\epsilon$ values larger than 1.0. Models estimated with $\epsilon > 1$ may significantly drift away from the true causal models.

\noindent\textbf{Remark:} Naturally, one might consider smaller values of $\epsilon$, such as $\epsilon = 0.1$. However, smaller $\epsilon$ values tend to induce only minimal changes in the causal models, making it more difficult to satisfy the constraints—particularly the interventional ones. In this work, the threshold parameter $\omega$ is set to 0.3, meaning that only edge coefficients greater than 0.3 are retained after thresholding. If there is only one causal path from a cause variable to an effect variable, and this path contains more than one edge (which is often the case), then having two edge coefficients each below 0.3 would result in a total causal effect less than $0.3 \times 0.3 = 0.09$, approximately 0.1. Only when the coefficients exceed 0.3 will they be preserved after thresholding, ensuring that the total causal effect of such a two-edge path is above 0.09. Moreover, in real-world settings, a causal effect from variable $i$ to $j$ is often not considered practically significant if $|T_{ij}| < 0.1$. This motivates our decision not to consider settings with $\epsilon < 0.25$, such as $\epsilon = 0.1$, in this study.

\noindent\textbf{Conclusion:} Based on the above analysis, we empirically conclude that $\epsilon = 0.25$ is a reasonable choice, offering the best accuracy performance.
\end{appendices}
\end{document}